\documentclass[preprint,12pt]{elsarticle}

\IfFileExists{lmodern.sty}{\usepackage{lmodern}}{}
\usepackage{amsmath}
\usepackage{amssymb}
\usepackage{amsfonts}
\usepackage{textcomp}
\usepackage{booktabs}
\usepackage{tabularx}
\usepackage{multirow}
\usepackage{rotating}
\usepackage{caption}
\usepackage{subcaption}
\usepackage{float}
\usepackage{placeins}
\usepackage{algorithm}
\usepackage{algpseudocode}
\usepackage{listings}
\usepackage{tikz}
\usetikzlibrary{shapes.geometric, arrows.meta, positioning, calc, fit, backgrounds}
\usepackage{hyperref}
\IfFileExists{microtype.sty}{\usepackage{microtype}}{}
\IfFileExists{newunicodechar.sty}{%
  \usepackage{newunicodechar}
  \newunicodechar{’}{\textquoteright}
  \newunicodechar{‘}{\textquoteleft}
  \newunicodechar{“}{\textquotedblleft}
  \newunicodechar{”}{\textquotedblright}
  \newunicodechar{–}{\textendash}
  \newunicodechar{—}{\textemdash}
}{}

\usepackage{etoolbox}
\makeatletter
\newcommand{\elsfrontpatch}[3]{%
  \patchcmd{#1}{#2}{#3}{\typeout{ELS: front-matter patch ok}}{%
    \typeout{ELS: WARNING: front-matter patch FAILED}}}
\elsfrontpatch{\pprintMaketitle}{\vskip36pt}{\vskip22pt}
\elsfrontpatch{\pprintMaketitle}{\vskip18pt}{\vskip14pt}
\elsfrontpatch{\pprintMaketitle}{\hrule\vskip12pt}{\hrule\vskip10pt}
\elsfrontpatch{\pprintMaketitle}{\hrule\vskip12pt}{\hrule\vskip10pt}
\elsfrontpatch{\pprintMaketitle}{\unvbox\absbox\par\vskip10pt}{\unvbox\absbox\par\vskip8pt}
\elsfrontpatch{\pprintMaketitle}{\unvbox\keybox\par\vskip10pt}{\unvbox\keybox\par\vskip8pt}
\elsfrontpatch{\MaketitleBox}{\vskip36pt}{\vskip22pt}
\elsfrontpatch{\MaketitleBox}{\vskip18pt}{\vskip14pt}
\elsfrontpatch{\MaketitleBox}{\hrule\vskip12pt}{\hrule\vskip10pt}
\elsfrontpatch{\MaketitleBox}{\hrule\vskip12pt}{\hrule\vskip10pt}
\elsfrontpatch{\MaketitleBox}{\unvbox\absbox\par\vskip10pt}{\unvbox\absbox\par\vskip8pt}
\elsfrontpatch{\MaketitleBox}{\unvbox\keybox\par\vskip10pt}{\unvbox\keybox\par\vskip8pt}
\elsfrontpatch{\keyword}{\raggedright}{\raggedright\interlinepenalty=10000\relax}
\makeatother

\begin{document}

\begin{frontmatter}

\title{A Hybrid RAG Framework for Deceptive Review Detection: Integrating RoBERTa-LoRA, FAISS-Based Retrieval, and LLM Reasoning}

\author[1]{Andrei Kostin\corref{cor1}}
\ead{ak000171@red.ujaen.es}

\author[2]{José Ángel Díaz García\fnref{fn1}}
\ead{joseangeldiazg@ugr.es}

\author[3]{Karel Gutiérrez-Batista\fnref{fn1}}
\ead{karel@decsai.ugr.es}

\cortext[cor1]{Corresponding author}
\fntext[fn1]{These authors contributed equally to this work.}

\affiliation[1]{
  organization={Department of Computer Science, University of Jaén},
  addressline={Campus Las Lagunillas},
  city={Jaén},
  postcode={23071},
  country={Spain}
}

\affiliation[2]{
  organization={Department of Computer Languages and Systems, University of Granada},
  addressline={Cortadura del Valle, s/n},
  city={Ceuta},
  postcode={51001},
  country={Spain}
}

\affiliation[3]{
  organization={Department of Computer Science and Artificial Intelligence, University of Granada},
  addressline={Cortadura del Valle, s/n},
  city={Ceuta},
  postcode={51001},
  country={Spain}
}

\enlargethispage{2\baselineskip}
\newpage

\begin{abstract}
The growth of deceptive online opinions poses a critical risk of misinformation and fraud. This paper proposes a Retrieval-Augmented Generation (RAG) system to automatically distinguish spam reviews. The architecture integrates RoBERTa with Low-Rank Adaptation (LoRA) classification, FAISS semantic vector retrieval, and a Large Language Model (LLM) for final reasoning. The system was evaluated across balanced and severely imbalanced real-world scenarios. Results demonstrate that five epochs of fine-tuning enable robust domain transfer, achieving an F1-score of $0.90$ in cross-domain evaluation, outperforming baselines (OpenAI Detector, $\text{F1} = 0.82$) and demonstrating competitive performance against specialized in-domain SOTA models (fakeRoBERTa, $\text{F1} = 0.97$). The proposed architecture provides evidence-backed explainability and statistically significant precision gains while maintaining comparable overall detection performance.
\end{abstract}


\end{frontmatter}

\enlargethispage{2\baselineskip}

\newpage
\section{Introduction}\label{sec1}

The rapid growth of e-commerce has radically changed customer habits, making online reviews one of the main factors influencing business reputation and consumer decisions in the digital world.  Nowadays, e-commerce platforms (for example, Amazon, Walmart) and special services  (such as Yelp, TripAdvisor) generate enormous amounts of data. By the end of 2022, Yelp's base had more than 265 million reviews \cite{Statista2023}. According to recent BrightLocal reports, 97\% of consumers read reviews before making a purchase, and 41\% of users consult them regularly \cite{Brightlocal2025}. The economic impact of these results is enormous; for example, a Harvard Business School study found that a one-star increase in a restaurant’s Yelp rating leads to a 5–9\% increase in revenue \cite{Luca2016}.

Finally, the significant financial impact of these numbers has led to the proliferation of fake reviews and an increased risk of misinformation. Some market participants follow a strategy known as astroturfing in which they use fraudulent reviews to artificially inflate their own reputation and to discredit competitors \cite {Aslam2019}. According to estimates by analytical resources, the proportion of fake reviews on major platforms such as Amazon can reach 30\% \cite{WiseReview2025}. This problem is getting worse with the appearance of modern large language models (LLMs) capable of mass-generating synthetic texts that are linguistically indistinguishable from those written by real people \cite{Papageorgiou2024}. As a result, traditional filters and manual moderation are losing their effectiveness, requiring the development of a new generation of intelligent automated systems.

Recent advances in expert systems have further demonstrated the effectiveness of integrating RAG with adaptive decision mechanisms for content verification. For instance, in \cite{lv2026lightweight}, the authors introduced SWIFT, a self-reflective knowledge-switching framework that dynamically triggers external retrieval only when parametric confidence drops below a threshold, significantly optimising inference efficiency. To mitigate the risk of hallucinated reasoning during arbitration, Yang et al. \cite{yang2026causality} proposed an agentic framework designed for evidence-inconsistency filtering before feeding context into LLMs. Additionally, Gao et al. \cite{gao2026sel} demonstrated with SEL-PMF that combining Parameter-Efficient Fine-Tuning with LLM-generated background evidence substantially enhances semantic representation in sparse textual contexts. Building upon these expert architectural principles, our work proposes a threshold-gated hybrid RAG system specifically engineered for deceptive online review detection under domain shift and class imbalance. Our research is guided by the following research questions: 


\begin{itemize}
    \item \textbf{RQ1}: How does retrieval-augmented generation affect the precision-recall trade-off in deceptive review detection?
    \item \textbf{RQ2}: To what extent does LoRA enable efficient cross-domain adaptation in deceptive review detection?
    \item \textbf{RQ3}: How robust is the proposed architecture under severe class imbalance?
    \item \textbf{RQ4}: Does retrieval-based reasoning improve explainability without degrading performance?
\end{itemize}

The contributions of this work are as follows:
{\sloppy
\begin{itemize}
\item A hybrid RAG architecture that combines a supervised classifier (RoBERTa-LoRA), a semantic retrieval module (FAISS) that returns both similarity scores and the actual text of similar past reviews, and a large language model (Llama 3.1) that uses all of this to make and explain the final decision.

\item A classifier that can adapt to a new domain and new class imbalance in only five additional training epochs, instead of retraining from scratch, and it was tested across three conditions: a balanced in-domain set, a severely imbalanced set, and a cross-domain set with AI-generated fakes.

\item An ablation study that measures how much each part of the system (classifier, retrieval, LLM reasoning) actually contributes to the final result. 

\item Giving the language model the actual text of retrieved similar reviews, not just distance scores, allows it to produce a short written justification for each decision, giving the system a level of explainability that the baseline classifier and simple k-NN approaches do not have.
\end{itemize}
}

From an operational perspective, automated moderation in modern e-commerce platforms cannot rely solely on black-box probabilistic scoring. Incorrectly penalizing legitimate seller reviews or missing sophisticated computer-generated spam introduces severe reputational and economic liabilities. To overcome this limitation, our hybrid architecture positions the RAG framework not merely as a binary classifier, but as an auditable expert decision-support system for online content moderation. By pairing high-precision vector retrieval with LLM arbitration, the system provides human safety teams with an inspectable evidence chain, transforming opaque machine learning predictions into a transparent, expert-assisted auditing workflow.

The rest of the paper is organised as follows. Section~\ref{sec2} provides an overview of related work in the field of spam detection and the application of deep learning. Section~\ref{sec3} describes the proposed system architecture in detail, including preprocessing steps, the configuration of the FAISS semantic search module, classifier operation, and the LLM inference mechanism. Section~\ref{sec4} details the LLM decision layer, prompt design, and confidence-threshold routing mechanism. Section~\ref{sec5} presents the experimental design, quantitative results, and ablation analysis. Section~\ref{sec6} discusses the findings in relation to the research questions. Finally, Section~\ref{sec7} summarises the study and outlines key directions for future work.

\section{Related works} \label{sec2}

Over the past decade, researchers have been actively studying the problem of detecting deceptive opinion spam. The evolution of approaches in this field can be divided into three key stages: the use of traditional machine learning, the application of deep learning models based on the transformer architecture, and the implementation of modern systems augmented with external search, such as RAG. 

\subsection{Classic Machine Learning}
Early research in the field of spam review detection relied on lexical analysis, n-grams, and psycholinguistic characteristics. A fundamental work in this area was the study by Ott et al., which introduced the first large-scale annotated dataset (the Deceptive Opinion Spam Corpus), consisting of 400 genuine and 400 fake hotel reviews \cite{Ott2011}. The fraudulent reviews were written by experts. The authors showed that humans can detect generated deception only at the level of random guessing (about 50\% accuracy), whereas automatic text-classification algorithms achieve nearly 90\% accuracy, confirming the critical need to automate this process.

On the other hand, Jindal and Liu conducted an analysis of 5.8 million reviews on the Amazon platform, focusing on identifying duplicate and near-duplicate reviews written by the same reviewers or about the same products as key indicators of spam activity \cite{Jindal2008}. Their approach demonstrated the importance of combining text analysis with the study of user behaviour patterns. However, classical methods (such as SVM, Logistic Regression, or Random Forest) based on TF-IDF features have limited ability to capture complex semantics and are vulnerable to modern generative models used by spammers.

\subsection{Deep Learning and Transformers}
With the development of neural network technologies, researchers have focused on deep learning models, such as CNNs and RNNs, and transformer-based architectures (e.g., BERT, RoBERTa), which are capable of extracting semantic patterns from text without the need for manual feature engineering \cite{Devlin2019, Liu2019}. In particular, Raza et al. applied BERT with progressive training and achieved strong results (95.3\% accuracy, 0.953 F1-score) on the WELFake dataset, which includes more than 72,000 articles, confirming the superiority of transformers over traditional machine learning \cite{Raza2025}.

At the same time, Papageorgiou et al. have noted in their research on the use of LLMs for dealing with fake content that modern neural networks play a dual role \cite{Papageorgiou2024}. In particular, they can act as high-quality synthetic spam or, on the other hand, serve as a powerful tool for detecting it. However, classic transformers are limited by the knowledge acquired during pretraining, tend to overfit under severe class imbalance, and function as ``black boxes'', providing no evidence to support their decisions. 

\subsection{RAG architecture}
To overcome the limitations of a lack of transparency and the external knowledge used by models, RAG architectures have been actively researched in recent years. The integration of RAG allows systems to dynamically retrieve relevant text examples from external knowledge bases and use them as additional context \cite{LopezJoya2025}. These factors improve the accuracy and explainability of the final responses in tasks requiring fact verification or fraudulent detection. This approach has already demonstrated high efficiency. In particular, Singal et al. developed a system that extracts evidence phrases from a knowledge base, which were then transferred to an LLM using context-aware training. This approach improved the performance of baseline models by 22\% with a minimal number of training examples \cite{Singal2024}. Also, the VeraCT Scan system successfully integrated semantic web search for fake news classification using GPT-4 Turbo and Llama-2 13B models, demonstrating state-of-the-art results on the BuzzFeedNews and PolitiFact datasets \cite{Niu2024}.

Despite the achieved progress of RAG systems in the field of news fact-checking, the application of this architecture to the detection of fake reviews remains largely unexplored. Unlike fact-checking, where falsity is determined by discrepancies with objective reality, fake reviews are identified through linguistic anomalies, triteness, and contextual inconsistencies.

\subsection{Existing Benchmarks and Baseline Models}
A growing challenge is the detection of AI-generated spam reviews. Salminen et al. created a cross-domain dataset of 40,000 Amazon reviews, half of which were generated using two language models, ULMFiT and GPT-2. In their study, baseline classifiers, such as NBSVM and the specialised fakeRoBERTa model, were trained and tested within a single domain (in-domain), achieving accuracies of up to 97\%, while humans were only able to solve the task 55\% of the time \cite{Salminen2022}. 

The proposed architecture fills an existing research gap by adapting the RAG methodology to the task of filtering online reviews. Unlike the baseline models mentioned in the last study, our system is designed for effective cross-domain transfer and resilience to class imbalance. The proposed system also offers high explainability and decision transparency through the extraction of semantic evidence.  

For clarity, Table \ref{tab:related_works_tfm} summarises the key studies in spam detection and content verification. As shown in this table, although existing transformer-based models achieve high accuracy in familiar domains, they lack explainability and require resource-intensive retraining. The proposed RAG architecture solves these issues.

\begin{table}[p]
\centering
\caption{Summary of studies in the field of spam and fake content detection (Related Work)}
\label{tab:related_works_tfm}
\scriptsize
\linespread{0.92}\selectfont
\renewcommand{\arraystretch}{1.0}
\setlength{\tabcolsep}{2pt}
\begin{tabularx}{\textwidth}{@{}>{\raggedright\arraybackslash}p{1.5cm}>{\raggedright\arraybackslash}X>{\raggedright\arraybackslash}p{2.05cm}>{\raggedright\arraybackslash}p{1.85cm}>{\raggedright\arraybackslash}X@{}}
\toprule
\textbf{Author} & \textbf{Description} & \textbf{Dataset} & \textbf{Outcomes} & \textbf{Research Gap} \\ 
\midrule

Ott et al. (2011) \cite{Ott2011} & 
First systematic study on deceptive reviews; utilisation of n-grams and psycholinguistic analysis & 
800 hotel reviews of 20 Chicago hotels (400 genuine, 400 fake) & 
Algorithm accuracy 89.8\%, whereas humans recognise fakes only 50\% of the time. & 
It needs to adapt to new domains; vulnerable to modern spammers. \\ 
\midrule
Jindal \& Liu (2008) \cite{Jindal2008} & 
The study categorises review spam into three types (untruthful opinions, brand-only reviews, ads). They used duplicate and near-duplicate review patterns to identify and analyse deceptive opinions with a logistic regression model & 
5.8 million reviews from the Amazon platform & 
Failure of text-only approaches (they obtained a poor AUC of only 
63\%) & 
It has a limitation in finding duplicates using a 2-gram-based content comparison with the Jaccard distance, and fails to capture subtle semantic patterns \\ 
\midrule
Raza et al. (2025) \cite{Raza2025} & 
Application of BERT architecture with progressive episode-based training strategy to capture linguistic nuances of fake news & 
WELFake dataset (72,134 articles). & 
Accuracy: 95.3\%, F1-score: 0.953. & 
The current model lacks transparent explainability in its decision-making. The WELFake dataset has potential imperfections\\ 
\midrule

Lopez-Joya et al. (2025) \cite{LopezJoya2025} & 
They proposed the VERIFAID RAG system for fake news detection. The system was built on an automatically generated, dynamically growing database. This design provides traceable evidence to support the final classification & 
VERIFAID-dataset (33,337 entries) & 
On benchmark datasets like GossipCop, CoAID, and PolitiFact, the VERIFAID framework achieved competitive F1-Macro and accuracy scores & 
The framework is designed specifically to detect general misinformation, fake news, and political or health-related rumours, not tailored to deceptive review spam \\ 
\midrule
Singal et al. (2024) \cite{Singal2024} & 
Automated fact-checking system based on RAG (using Mixtral 8x22B) and in-context learning (Few-Shot ICL) to provide evidence-backed predictions & 
AVeriTeC dataset (4,568 real-world claims) & 
Improved baseline metrics by 22\% with a minimal number of training examples & 
The extracted QA evidence is susceptible to LLM hallucinations. Tuned exclusively for fact-checking, not for review analysis \\ 
\midrule
Niu et al. (2024) \cite{Niu2024} & 
The VeraCT RAG system integrating web search and optimised query generation for transparent decision reasoning using GPT-4 and Llama-2 & 
BuzzFeedNews, Fakenewsnet, LLMFake, PolitiFact-Snopes-2024 and FakeNews2024 datasets & 
Achieved state-of-the-art results in news verification tasks & 
Complex web search architecture is not adapted for isolated review platforms
\end{tabularx}
\end{table}
\FloatBarrier

\section{System architecture} \label{sec3}
\subsection{Overview}
The proposed system for fake review detection is based on a three-level hybrid RAG architecture. As compared with traditional approaches, our architecture combines three independent prediction sources: a supervised classifier based on RoBERTa and LoRA adapters, which gives a primary spam probability; a semantic retrieval module based on a FAISS index, which extracts some examples as evidence with retrieval confidence metrics; and, finally, a decision layer based on an LLM (Llama-3.1-8B) for final decision and logical inference based on all collected signals from the two previous system layers.

\begin{figure*}[!t]
\centering
\resizebox{0.98\textwidth}{!}{
\begin{tikzpicture}[
    >=Stealth,
    font=\small\sffamily,
    box/.style={draw=black!75, fill=gray!8, rounded corners=3pt, align=center, inner sep=6pt, line width=0.8pt},
    proc/.style={box, fill=blue!8, draw=blue!65},
    clf/.style={box, fill=orange!10, draw=orange!65},
    rag/.style={box, fill=teal!10, draw=teal!65},
    decision/.style={diamond, draw=red!65, fill=red!8, aspect=2.2, inner sep=2pt, line width=0.8pt, align=center},
    data/.style={cylinder, draw=gray!75, fill=gray!15, shape border rotate=90, aspect=0.25, inner sep=4pt, align=center},
    subframe/.style={draw=black!35, dashed, rounded corners=5pt, inner sep=20pt, fill=black!2},
    framelabel/.style={anchor=south west, font=\footnotesize\bfseries, text=black!80}
]

\node[data] (train) at (0,-0.5)
    {Training Data\\(OpSpam + Yelp\\Labeled Reviews)};
\node[data] (vector) at (0,-3.2)
    {Vector Knowledge Base \\(YelpZIP + YelpNYC)};

\node[proc] (embed) at (6.5,-1.6)
    {Embedding Extraction\\(RoBERTa-base)};

\node[clf]  (classifier) at (14.7,0)
    {RoBERTa-LoRA\\Classifier};
\node[data] (faiss) at (14.7,-3.2)
    {\\FAISS Index\\};

\coordinate (split) at (9.8,-1.6);

\draw[->, thick] (train.east) -| (embed.north);
\draw[->, thick] (vector.east) -| (embed.south);
\draw[thick] (embed.east) -- (split);
\draw[->, thick] (split) |- node[above, pos=0.78] {train} (classifier.west);
\draw[->, thick] (split) |- node[below, pos=0.78] {index} (faiss.west);

\node[box] (input) at (0,-9.5) {Unseen Review $x$};
\node[proc] (query_embed) at (4.5,-9.5)
    {Feature Extractor\\$\mathbf{e}(x)$ + RoBERTa-LoRA};
\node[decision] (check) at (9.8,-9.5) {$p_{\text{base}}\in [\tau_{low}, \tau_{high}]$?};

\node[rag] (retrieve) at (14.7,-8)
    {$k$-NN Retrieval (FAISS)\\closest distance,\\fake ratio,\\ top-3 examples};
\node[rag] (llm) at (22,-8) {LLM Arbitration};
\node[box, fill=teal!20, draw=teal!75] (output) at (26,-8)
    {Final Prediction $\hat{y}$\\+ Justification};

\node[box] (direct) at (14.7,-11.2) {Direct Label\\$\hat{y} = \mathbb{I}(p_{\text{base}} \ge 0.5)$};

\draw[->, thick] (input) -- (query_embed);
\draw[->, thick] (query_embed) -- node[above] {$p_{\text{base}}$} (check);
\draw[->, thick] (check.north) |- node[pos=0.75, above] {Yes (Uncertain)} (retrieve.west);
\draw[->, thick] (check.south) |- node[pos=0.75, below] {No (Confident)} (direct.west);
\draw[->, thick] (retrieve) -- node[above] {Retrieval Summary} (llm);
\draw[->, thick] (llm) -- (output);
\draw[->, thick] (direct.east) -| (output.south);

\draw[->, thick, dashed] (faiss.south) -- (retrieve.north);
\coordinate (w1) at (16.5,0);
\coordinate (w2) at (16.5,-5.7);
\coordinate (w3) at (4.5,-5.7);
\draw[->, thick, dashed] (classifier.east) -- (w1) -- (w2)
    -- node[above] {trained LoRA weights} (w3)
    -- (query_embed.north);
\node[data] (newdom) at (12,-15.2)
    {New-Domain Data\\(other product\\categories\\or imbalance)};
\node[clf] (adapt) at (4.5,-15.2)
    {LoRA Fine-Tuning\\(5 epochs)};

\draw[->, thick] (newdom) -- node[above] {domain shift} (adapt);
\draw[->, thick, dashed] (adapt.north) -- node[pos=0.7, right=4pt, align=left] {updated\\LoRA weights} (query_embed.south);
\begin{scope}[on background layer]
    \node[subframe, fit=(vector)(train)(embed)(classifier)(faiss),
          label={[framelabel]north west:(A) Training \& Indexing Phase}] (frameA) {};

    \node[subframe, inner sep=24pt,
          fit=(input)(query_embed)(check)(direct)(retrieve)(llm)(output)(newdom)(adapt),
          label={[framelabel]north west:(B) Online Inference}] (frameB) {};

    \node[subframe, draw=orange!65, fill=orange!5, inner sep=10pt, fit=(newdom)(adapt),
          label={[framelabel, anchor=south, text=orange!70!black]north :(C) Domain Adaptation}] (frameC) {};
\end{scope}

\end{tikzpicture}
}
\caption{Diagram of the proposed Hybrid RAG architecture.}
\label{fig:architecture}
\end{figure*}
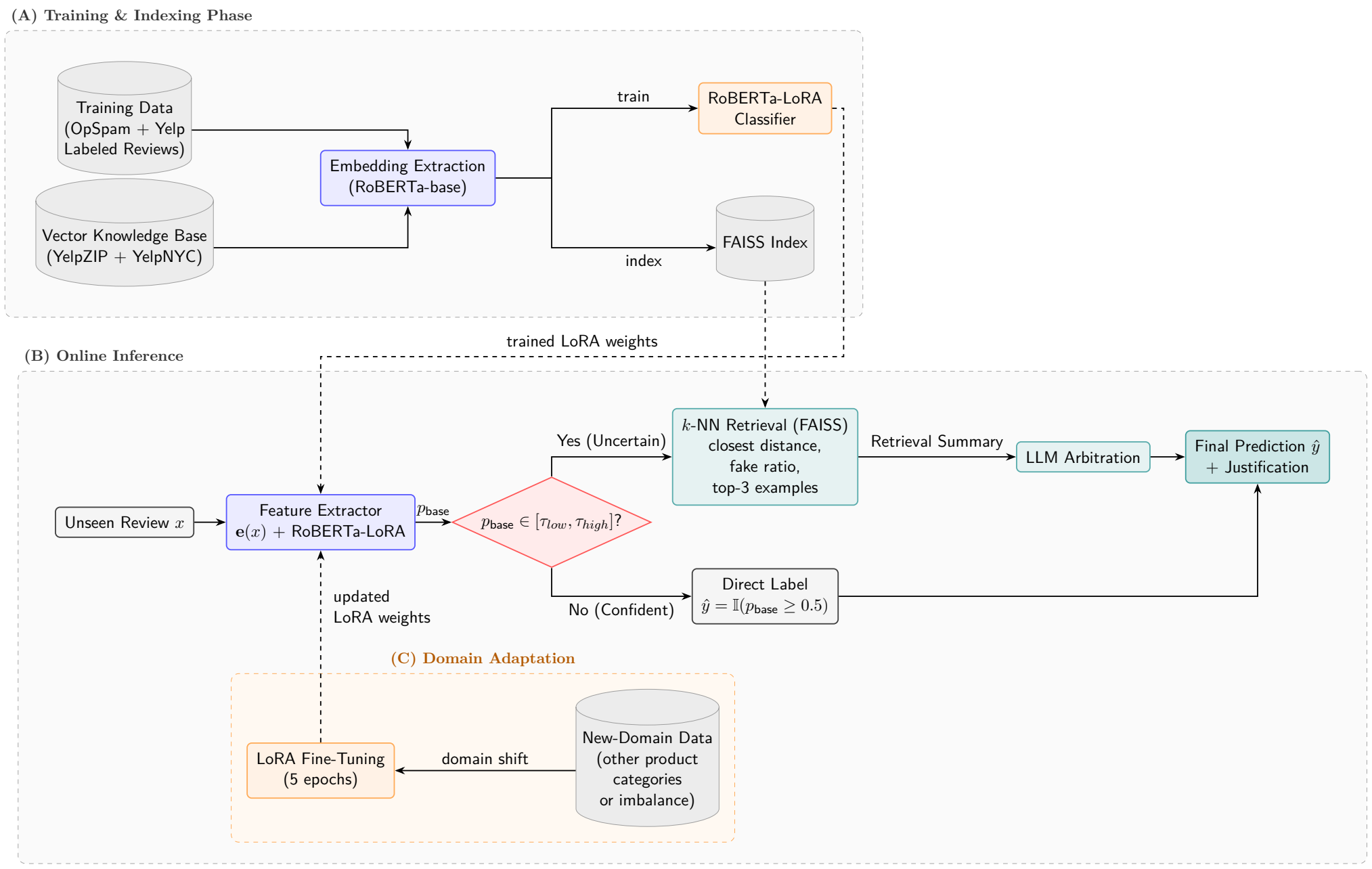
\FloatBarrier

As shown in Fig. \ref{fig:architecture}, the proposed architecture has two pipelines plus a domain-adaptation step. Firstly, the training pipeline includes data preprocessing, embedding extraction, building a FAISS index based on labelled Yelp HoReCa reviews, and training the RoBERTa-LoRA classifier on annotated data. Second, the inference pipeline for new data vectorises inputs, obtains a probability from the classifier, retrieves confidence metrics for nearest neighbours from the knowledge-base FAISS index, and lets the LLM make the final decision.

\subsection{Datasets} \label{subsec3.2}
In this study, different datasets were used for classifier training, knowledge base construction, and diverse system evaluation. Every dataset plays a strictly defined function in the architecture. The general descriptions and roles of all datasets are presented in Table \ref{tab:datasets_summary}.

\begin{table}[htbp]
\centering
\small
\caption{Summary of datasets and their roles in the proposed system}
\label{tab:datasets_summary}
\begin{tabular}{@{}>{\raggedright\arraybackslash}p{0.22\textwidth}>{\raggedright\arraybackslash}p{0.22\textwidth}>{\raggedright\arraybackslash}p{0.46\textwidth}@{}}
\toprule
\textbf{Dataset} & \textbf{Role in System} & \textbf{Size and Details} \\
\midrule
Deceptive Opinion Spam Corpus v1.4 \cite{Ott2011} & Classifier training \newline(first dataset\textsuperscript{1}) & 1,600 hotel reviews; 800 genuine / 800 fake. Split 70/30 (1,120 train / 480 validation). \\
\addlinespace
Yelp Fake Review Dataset \cite{df_yelp}  & Classifier training \newline(second dataset\textsuperscript{1})  & 9,929 train / 2,483 test instances. Restaurant reviews with realistic class skew based on Yelp's filter. \\
\addlinespace
YelpNYC + YelpZIP \cite{df_faiss} & FAISS Knowledge Base & 768,683 labelled reviews by Yelp platform filter (86\% genuine / 14\% fake). Used exclusively to build a large-scale semantic index. \\
\addlinespace
In-Domain Data  (Test set split) & Evaluation \newline(Exp. 1: Baseline) & 1,309 instances from the combined training data. Stratified split (70\% genuine / 30\% fake) with fixed seed. \\
\addlinespace

Hotel Reviews \cite{df_hotelReviews} & Evaluation \newline(Exp. 2 with imbalance) & Sample of 3,000 reviews. Severe class imbalance: 2,700 genuine (90\%) / 300 fake (10\%). \\
\addlinespace
Cross-domain Reviews \cite{Salminen2022} & Evaluation \newline(Exp. 3 with cross-domain reviews) & Sample of 3,000 multi-domain reviews (50\% genuine / 50\% LLM-generated fakes). \\
\bottomrule
\end{tabular}
\par\vspace{4pt}
\begin{minipage}{\textwidth}
\footnotesize\raggedright
\textsuperscript{1}Note: After preprocessing, the annotated datasets (Opinion Spam and Yelp) were combined into a single training corpus with a 70/30 stratified split to preserve class proportions.\par
\end{minipage}
\end{table}

\subsection{Preprocessing}

During preprocessing of the selected datasets, a unified data normalisation pipeline was implemented. Specifically, this includes basic cleaning of noise symbols and the removal of rows containing empty opinion text and very short reviews that could impair the performance of the spam detection model. Likewise, words were not converted to lowercase to avoid losing potential context in fraudulent or genuine opinions—such as the use of Caps Lock to express emotions.

The decision to preserve the case of text characters is crucial from a linguistic perspective of a review, since the use of uppercase characters (``AMAZING'', ``BEST EVER’’) may be related to the underlying context of the review as a whole and must be analyzed by the system.

The specific stages of data preprocessing include:
\begin{enumerate}
    \item Removal of HTML tags, special characters, and null values.
    \item Normalisation of whitespace: replace multiple spaces, tabs, and line breaks with a single space.
    \item Removal of text shorter than 30 characters, as it lacks sufficient linguistic context for initial classification.
    \item Removal of duplicate reviews to avoid overfitting on repetitive examples.
    \item Label normalisation.
\end{enumerate}

To train the supervised classifier, the Opinion Spam Corpus v1.4 and Yelp Fake Reviews Dataset were preprocessed and merged into a single training corpus. This combined dataset was partitioned using a stratified 70\%/30\% train-validation split. Stratification ensured that the class ratios are maintained in both sets, which is crucial for an accurate comparison of the results. 

For the vector knowledge base, the YelpNYC and YelpZIP were combined into a unified large-scale corpus. The text duplicates were removed from the combined corpus to maintain index integrity, prevent memory redundancy, and eliminate bias during FAISS k-NN semantic retrieval.

\subsection{Feature Extraction}
The next step after data preprocessing is to obtain a vector representation of the text. To do this, the pre-trained RoBERTa (Robustly Optimised BERT Pretraining Approach) model was used, which is better suited for detecting fake opinions because it is more robust to context preservation and variability. It also significantly improves results compared to more basic methods, such as TF-IDF, as will be demonstrated later.

Specifically, 768-dimensional embeddings from the \texttt{roberta-base} model were extracted from the merged training dataset (Yelp + Deceptive Opinion Spam Corpus) to train the classifier, and were also obtained from the large combined corpus (YelpNYC + YelpZIP) to build the FAISS index, which will be described in the following section.

RoBERTa was selected as the base encoder model for several reasons. First, compared to the original BERT, RoBERTa was trained for a longer period and with more data, which improves the quality of contextual representations \cite{Liu2019}. Additionally, RoBERTa uses dynamic masking during training, which increases its ability to adapt to the input text.

\subsection{Supervised Classifier (RoBERTa-LoRA)}
The initial review classification is performed using a RoBERTa model fine-tuned with the Parameter-Efficient Fine-Tuning (PEFT) library \cite{Xu2026parameter}. The Low-Rank Adaptation (LoRA) mechanism was applied to reduce the computational costs of retraining a large model \cite{LoRa_paper}. LoRA freezes the model's initial weights $W_0$ and applies updates via low-rank decomposition matrices ($A$ and $B$). Mathematically, the forward pass is expressed as:
\begin{equation}
        h = W_0 x + \Delta W x = W_0 x + BAx
\end{equation}
where, $B \in \mathbb{R}^{d \times r}$, $A \in \mathbb{R}^{r \times k}$, and $r \ll min(d,k)$.

The adapter’s hyperparameters were set as follows: rank $r=8$, $\alpha=16$, and $dropout=0.1$. It was trained for 10 epochs using the AdamW optimiser with weight decay regularisation to prevent overfitting. A CustomLossTrainer module for fine-tuning was developed with a weighted cross-entropy loss function to effectively handle imbalanced real-world data. The class weights ($w_k$) were calculated dynamically using the following formula:
\begin{equation}
        w_k=\frac{N}{K \cdot n_k}
\end{equation}
where, $N$ is the total number of examples, $K$ is the number of classes, and $n_k$ is the number of examples in class $k$. For example, with a 90/10 imbalance, the model penalises an error on the minority class about 9 times more severely.

A weighted Cross-Entropy Loss function, mathematically, is calculated as:

\begin{equation} \mathcal{L}_{\text{weighted}} = -\frac{1}{N} \sum_{j=1}^{N} \left[ w_1\, t_j \log(p_j) + w_0\,(1-t_j)\log(1-p_j) \right] 
\end{equation}

where $N$ is the total number of objects in the sample, $t_j$ is the target label for binary classification, $p_j$ is the model's prediction that the opinion belongs to class 1 (false), and $w_1$ and $w_0$ are the weights assigned to fake and genuine reviews respectively.

During research, a linear interpolation of the classifier probabilities and the k-NN probabilities from FAISS ($P_{final} = \lambda P_{clf} + (1-\lambda) P_{kNN}$) was tested, but this approach was rejected because it did not outperform the standalone LoRA classifier in the F1 metric.

\subsection{Retrieval Module (FAISS)}
To provide the system with an evidence-search mechanism, a verified vector database from the combined Yelp dataset (YelpNYC + YelpZIP) was constructed. This base was implemented using the FAISS (Facebook AI Similarity Search) library \cite{Johnson2019FAISS}. The IndexFlatL2 index was used, which performs an exact search for nearest neighbours based on Euclidean distance in $\mathbb{R}^{768}$ dimensions. Embeddings were generated in batches of 512 examples to optimise memory usage.

During inference, after receiving the input data, the system searches for the top-k nearest neighbours. For each query, the system retrieves $k=20$ nearest neighbours. This value was selected empirically to balance stability and retrieval relevance. Too small values ($k=3$) give unstable results, while too large values ($k=100$) increase the number of irrelevant outliers and degrade the retrieval signal. Since the knowledge base is fully labelled, the retrieval module extracts quantitative similarity metrics and qualitative textual evidence, some of which are formatted into the retrieval summary for the LLM:
\begin{itemize}
    \item \textbf{Distance to the nearest neighbor} (\texttt{top1\_score}): the minimum Euclidean distance $d_{min}(p,q) = \min(\sqrt{\sum_{k=1}^{n} (p_k - q_k)^2})$. A low value indicates that there is an almost identical example in the knowledge base, which increases the reliability of the contextual signal.
    \item \textbf{Mean distance to neighbours} (\texttt{mean\_topk\_score}): the average distance of the k neighbours found, $d_{mean} = \frac{1}{k}\sum_{i=1}^{k} d_i$. It demonstrates the overall density of the neighbourhood through a query in the knowledge base space.
    \item \textbf{Difference between the average distance and the nearest neighbor} (\texttt{score\_gap}): The difference between the two previously calculated measures is $\Delta d = d_{mean} - d_{min}$. A large difference indicates that the first neighbour is significantly closer than the others. It means that the query has entered a sparse area with a similar example. A low result indicates a dense cluster of similar examples.
    \item \textbf{Deceptive Ratio in Top-k Neighbors}
    (\texttt{fake\_ratio}): the proportion of verified deceptive reviews among the k retrieved neighbours, which is calculated as $\text{Fake Ratio} = \frac{1}{k} \sum_{i=1}^{k} y_i, \quad y_i \in \{0, 1\}$, where $y_i = 1$ represents a fake review and $y_i = 0$ represents a genuine review. A ratio close to $1.0$ signals that the target review falls into a dense cluster of spam opinions.
    \item \textbf{Top-3 Similar Textual Examples}: The raw review texts of the 3 closest neighbours to provide the LLM with direct contextual evidence for comparative analysis. 
\end{itemize}

\subsection{Inference and Domain Adaptation} 
One of the system’s key innovations is the domain adaptation during the inference layer. When the system faces input data from a new domain (in our case, a transition occurs from hotel/restaurant reviews to other product categories), the RoBERTa-LoRA classifier undergoes additional retraining for 5 epochs. Thanks to LoRA’s computational efficiency, this step allows the model to quickly adapt to a new feature distribution and specific class imbalances.

\section{LLM Decision Layer} \label{sec4}
The locally deployed large language model Llama-3.1-8B-Instruct (quantised to 8-bit, Q8) was used as the final decision-making layer \cite{llama}. We chose this model for its strong instruction-following capabilities and high reasoning quality, while keeping it computationally efficient for local inference. The Llama model with 8 billion parameters can correctly handle strictly structured prompts and return responses in a specified format. We developed a specialised two-part prompt to ensure strict output formatting and minimise hallucinations. This approach includes a \textbf{System Message} and a \textbf{Human Template}.

The prompt engineering process for the system message included several iterations. In the initial phase, the model’s specific function and output format were specified. Based on the results of the classification error analysis, clear rules were added to the prompt that prohibit the model from interpreting the positivity, length, or level of detail of the text as indications of fakeness. Finally, a rigorous output schema with specified fields was introduced, ensuring a deterministic syntactic analysis of the model’s responses. The text of the system message is provided below:

\begin{lstlisting}
You are an expert in deceptive opinion spam detection.
Classify the target review:
- 1 = fake/spam
- 0 = real
Rules:
- Final decision is yours.
- Classifier result is supportive, not decisive.
- Retrieval is highly supportive. 
- Generic, templated, promotional, repetitive text -> LABEL 1.
- Grounded, situational, imperfect, experience-based text -> LABEL 0.
- Do not treat positivity, length, detail, or emotion as fake by itself.
- Output only in the exact format below.
Output exactly:
LABEL: 0|1
REASONING: [Your reasoning, explicitly mentioning the retrieved examples, fake ratio, and distance metric, and prioritizing them over the classifier result]
\end{lstlisting}

The prompt positions the language model as an expert in spam detection. The algorithm instructs the LLM to summarise the target review's textual characteristics with external evidence, treating the classifier result as supportive and the retrieval metrics as highly supportive. This was done to ensure that noise in the extracted data would not distort the model’s internal logic. The strict output format makes it easy to retrieve and analyse responses from the LLM. Clear rules guided the model to identify formulaic or promotional phrasing as spam, while treating experience-grounded details as genuine. The strict output requires a binary label, and a brief explanation ensures deterministic response parsing and explainable decision-making.

The Human Template is used to dynamically transmit the extracted data (FAISS retrieval summary including distance metric, neighbour fake ratio, and top-3 closest examples from the verified vector database, and classifier probability), as well as the review text for each specific query. Its text is provided below:
\\
\begin{lstlisting}
RETRIEVAL SUMMARY:
- Closest distance: {top1_score} (Lower is more similar to memory)
- Fake Ratio in Top-{neighbor_count} neighbors: {fake_ratio}     
(1.0 = all closest are fake, 0.0 = all real)
TOP 3 SIMILAR EXAMPLES FROM VERIFIED DATABASE:
1. {retrieval_example_1}
2. {retrieval_example_2}
3. {retrieval_example_3}
CLASSIFIER RESULTS:
- Probability of spam: {p_clf}
TARGET REVIEW TEXT:
{text}
\end{lstlisting}

One of the system's engineering solutions is a confidence threshold that lets the system skip the LLM when the classifier shows high confidence in its prediction. For a review $x$ with an embedding $e$, the RoBERTa-LoRA classifier gives a spam probability $p_{\text{base}} = P(\text{spam} \mid x)$. Two fixed thresholds are used:
$\tau_{low} = 0.3$ and $\tau_{high} = 0.7$.

\begin{itemize}
    \item If $p_{\text{base}} < 0.3$ or $p_{\text{base}} > 0.7$, the classifier is
    considered confident. The LLM is not called, and the system returns the
    classifier's label, $\hat{y}_{\text{base}} = \mathbb{I}(p_{\text{base}} \geq 0.5)$.
    \item If $0.3 \leq p_{\text{base}} \leq 0.7$, the review falls in the
    uncertainty window. The system retrieves the evidence from FAISS and
    sends it to the LLM, which makes the final decision.
\end{itemize}

In other words, the LLM only sees the reviews where the classifier is close to a coin flip. This routing logic is shown in Algorithm~\ref{alg:threshold_routing}.
The same thresholds are used in all experiments. The values $0.3$ and $0.7$ were set symmetrically around $0.5$ as a simple
choice and were not tuned. In our experiments, this window contained about
48\% of the zero-shot cross-domain reviews and about 21\% after fine-tuning
(Section~\ref{sec5}). Threshold tuning is left for future work.

This approach achieves two primary objectives:
\begin{itemize}
    \item \textbf{Computational Resource Efficiency}: By bypassing the generative layer for high-confidence samples, the mechanism drastically reduces GPU compute cycles, memory consumption, and overall inference latency.
    \item \textbf{Environmental Sustainability}: By avoiding unnecessary LLM forward passes on reviews that the classifier is confident about, the system significantly reduces energy footprint.
\end{itemize}

\begin{algorithm}[H]
\caption{Confidence-Threshold LLM Routing}\label{alg:threshold_routing}
\begin{algorithmic}[1]
\Require Review text $x$, Embedding $e$, Base probability $p_{\text{base}}$, Thresholds $[\tau_{\text{low}}, \tau_{\text{high}}]$, FAISS index $\mathcal{I}$, Neighbors $k$
\Ensure Final prediction $\hat{y} \in \{0, 1\}$

\State $\hat{y}_{\text{base}} \Leftarrow \mathbb{I}(p_{\text{base}} \ge 0.5)$

\If{$p_{\text{base}} < \tau_{\text{low}} \lor p_{\text{base}} > \tau_{\text{high}}$} \Comment{High confidence region: bypass LLM}
    \State \Return $\hat{y}_{\text{base}}$
\Else \Comment{Uncertainty region: $\tau_{\text{low}} \le p_{\text{base}} \le \tau_{\text{high}}$}
    \If{$\mathcal{I} \neq \emptyset$}
        \State $\mathcal{C} \Leftarrow \text{FAISS\_Search}(\mathcal{I}, e, k)$ \Comment{Retrieve similarity stats \& top-3 examples}
    \Else
        \State $\mathcal{C} \Leftarrow \emptyset$
    \EndIf
    \State $R \Leftarrow \text{Invoke\_LLM}(x, p_{\text{base}}, \hat{y}_{\text{base}}, \mathcal{C})$
    \State $\hat{y}_{\text{llm}} \Leftarrow \text{ExtractLabel}(R)$
    \If{$\hat{y}_{\text{llm}} \neq \text{None}$}
        \State \Return $\hat{y}_{\text{llm}}$
    \Else
        \State \Return $\hat{y}_{\text{base}}$ \Comment{Parsing error of llm response}
    \EndIf
\EndIf
\end{algorithmic}
\end{algorithm}

\section{Results}\label{sec5}
\subsection{Experimental Scenarios}
As described in detail in Section \ref{subsec3.2}, three experimental scenarios were developed based on the prepared datasets to  evaluate the RAG system:
\begin{enumerate}
    \item Baseline Scenario: Testing on a familiar domain (hotels and restaurants) with a 70/30 stratified imbalance.
    \item Extreme Imbalance Scenario: Testing on an external set of Hotel Reviews with a realistic class imbalance (90/10). 
    \item Cross-Domain AI-Spam Scenario: Testing on a balanced (50/50) Cross-domain Reviews dataset, where the fake reviews are generated by artificial intelligence. It was a challenge for the system’s generalisation ability.
\end{enumerate}

\subsection{Evaluation Metrics}
Given the nature of the task, the model was evaluated using the following metrics: Accuracy, Precision, Recall, F1-score, and PR-AUC. Particular attention was paid to Precision and F1-score on the fake class. In content moderation, a false positive causes direct reputational and commercial harm to sellers, so Precision determines the practical applicability of the system. F1-score as the balanced measure shows whether the gain in Precision is achieved without a significant loss of overall detection quality.

\subsection{Classifier Experiments}
Before development of the final RAG pipeline, a comparison of approaches for initial classification was conducted, ranging from baseline models (TF-IDF) to fine-tuned contextual embeddings (RoBERTa-LoRA).  

As summarised in Table~\ref{tab:comparison_clfs_models}, experiments showed that bag-of-words baselines have noticeable limitations. While TF-IDF with Logistic Regression and Linear SVM achieved moderate F1-scores ($0.6024$ and $0.6040$, respectively), the complementary Naive Bayes baseline suffered from severe majority class bias, failing to detect fake reviews (Recall of $0.0897$ and F1-score of $0.1639$). 

RoBERTa embedding representations with traditional classifiers improved detection sensitivity in balanced settings; for example, Logistic Regression achieved a Recall of $0.6923$, but its low precision ($0.5590$) degraded the overall PR-AUC to $0.6617$. Tree-based ensembles (Random Forest and LightGBM) similarly struggled with the dense representation space under class skew.

In contrast, parameter-efficient fine-tuning of RoBERTa with Low-Rank Adaptation (LoRA) and a weighted cross-entropy loss function demonstrated significant performance across all metrics. This approach established an optimal balance on the deceptive class: an Accuracy of $0.8130$, Precision of $0.6862$, Recall of $0.6880$, and an F1-score of $0.6871$.

\begin{table}[htbp]
\centering
\small
\setlength{\tabcolsep}{4pt}
\caption{Comparison of different classification models on the test dataset}
\label{tab:comparison_clfs_models}
\begin{tabularx}{\textwidth}{@{}>{\raggedright\arraybackslash}Xccccc@{}}
\toprule
\textbf{Model} & \textbf{Acc} & \textbf{F1-Score} & \textbf{Prec} & \textbf{Rec} & \textbf{PR-AUC} \\
\midrule
TF-IDF + LogReg & 0.7701 & 0.6024 & 0.6213 & 0.5846 & 0.7030 \\
TF-IDF + LinearSVM & 0.7907 & 0.6040 & 0.6921 & 0.5359 & \textbf{0.7172} \\
TF-IDF + SVM & 0.8067 & 0.5886 & 0.8044 & 0.4641 & 0.7076 \\
TF-IDF + Naive Bayes & 0.7273 & 0.1639 & \textbf{0.9459} & 0.0897 & 0.6327 \\

RoBERTa + LogReg & 0.7456 & 0.6186 & 0.5590 & \textbf{0.6923} & 0.6617 \\
RoBERTa + LinearSVM & 0.7441 & 0.6109 & 0.5584 & 0.6744 & 0.6382 \\
RoBERTa + SVM & 0.6914 & 0.5258 & 0.4848 & 0.5744 & 0.6019 \\
RoBERTa + Random Forest & 0.7823 & 0.4991 & 0.7933 & 0.3641 & 0.6629 \\
RoBERTa + LightGBM & 0.7807 & 0.5423 & 0.7173 & 0.4359 & 0.6798 \\

\textbf{RoBERTa-LoRA} & \textbf{0.8130} & \textbf{0.6871} & 0.6862 & {0.6880} & {0.6918} \\
\bottomrule
\end{tabularx}
\end{table}
\FloatBarrier

\subsection{RAG System Evaluation}
The final evaluation of the complete RAG architecture was conducted on external datasets (the Hotel Reviews dataset and a cross-domain dataset) in two modes: zero-shot (without retraining) and with lightweight retraining of the classifier for 5 epochs. 

For the 5-epoch domain adaptation experiments, each target dataset was split into a 60\% adaptation set and a 40\% test set for evaluation using stratified sampling and a fixed random seed. The RoBERTa-LoRA classifier underwent lightweight fine-tuning on the adaptation split for 5 epochs, after which the full RAG pipeline was evaluated strictly on the unseen test data.

The summary results for all scenarios are presented in Table \ref{tab:rag_results}.

\begin{table}[htbp]
\centering
\small
\setlength{\tabcolsep}{4pt}
\caption{Performance comparison between baseline RoBERTa-LoRA and the full RAG pipeline across all evaluation scenarios}
\label{tab:rag_results}
\begin{tabularx}{\textwidth}{@{}>{\raggedright\arraybackslash}Xlcccc@{}}
\toprule
\textbf{Dataset} & \textbf{Setup} & \textbf{Acc}  & \textbf{Prec} & \textbf{Rec} & \textbf{F1-Score}\\
\midrule
\multicolumn{6}{l}{\textit{In-Domain Baseline (70/30)}} \\
& RoBERTa-LoRA only & 0.8130 & 0.6862 & \textbf{0.6880} & \textbf{0.6871} \\
& Full RAG Pipeline & \textbf{0.8237} & \textbf{0.7424} & 0.6266 & 0.6796 \\
\midrule
\multicolumn{6}{l}{\textit{Hotel Reviews (90/10, zero-shot)}} \\
& RoBERTa-LoRA only & 0.6712 & 0.1684 & \textbf{0.6007}  & 0.2631  \\
& Full RAG Pipeline & \textbf{0.7656} & \textbf{0.1967} & 0.4539 & \textbf{0.2745} \\
\midrule
\multicolumn{6}{l}{\textit{Hotel Reviews (90/10, 5-epoch adaptation)}} \\
& RoBERTa-LoRA only & 0.7083 & 0.2132 & \textbf{0.6136}  & 0.3164 \\
& Full RAG Pipeline & \textbf{0.8592} & \textbf{0.3520} & 0.3333  & \textbf{0.3424} \\
\midrule
\multicolumn{6}{l}{\textit{Cross-Domain Reviews (50/50, zero-shot)}} \\
& RoBERTa-LoRA only & 0.5790 & 0.5613 & \textbf{0.6819}  & \textbf{0.6158} \\
& Full RAG Pipeline & \textbf{0.6300} & \textbf{0.6391} & 0.5788 & 0.6075 \\
\midrule
\multicolumn{6}{l}{\textit{Cross-Domain Reviews (50/50, 5-epoch adaptation)}} \\
& RoBERTa-LoRA only & 0.8300 & 0.7651 & \textbf{0.9283} & 0.8389 \\
& Full RAG Pipeline & \textbf{0.8433} & \textbf{0.8067} & 0.8829 & \textbf{0.8431} \\
\bottomrule
\end{tabularx}
\end{table}
\FloatBarrier

The presented experimental comparison reveals the key patterns of the proposed RAG architecture across various domains and under different class distribution configurations. In all tested scenarios, the full RAG pipeline improves precision and accuracy compared to the baseline RoBERTa-LoRA classifier. Specifically, on the In-Domain dataset, precision increases from $0.6862$ to $0.7424$; on the fine-tuned Cross-Domain dataset, it increases from $0.7651$ to $0.8067$; and on the fine-tuned Hotel Reviews dataset, it rises from $0.2132$ to $0.3520$. This consistent improvement indicates that combining retrieved examples with LLM reasoning acts as a highly effective verification filter to eliminate false alarms.

This improvement in decision confidence is accompanied by a noticeable trade-off between precision and recall. However, this compromise is advantageous for practical e-commerce deployments, where falsely flagging genuine user reviews causes greater reputational and commercial harm than missing some fake reviews.

Furthermore, the results demonstrate the importance of a lightweight and efficient adaptation. Evaluating the framework in a pure zero-shot setting across new domains yields modest baseline results, whereas applying just five additional epochs of LoRA fine-tuning boosts performance, elevating Cross-Domain accuracy from $0.6300$ to $0.8433$ and F1-score to $0.8431$. Similarly, under severe class imbalance in the Hotel Reviews case, the 5-epoch adaptation improves overall accuracy from $0.7083$ to $0.8592$, increases precision from $0.2132$ to $0.3520$, and improves the F1-score from $0.3164$ to $0.3424$.

\subsubsection{Statistical Significance Analysis}
To ensure that the performance changes resulting from the use of the RAG architecture are statistically significant and not due to random sampling,  a paired bootstrap resampling test is performed between the Full RAG system and RoBERTa-LoRA with a $95\%$ confidence level ($N = 2,000$ iterations). The statistical evaluation was executed across the three primary benchmark scenarios: In-Domain set, Fine-Tuned Hotel Reviews, and Fine-Tuned Cross-Domain. Table \ref{tab:bootstrap_grouped} shows the results of the statistical analysis. 

\begin{table}[t]
  \centering
  \scriptsize
  \caption{Paired Bootstrap Significance Analysis ($N = 2{,}000$). Asterisks ($^*$) denote statistically significant differences in favour of our proposal, while double asterisks ($^{**}$) mean significant differences in favour of RoBERTa-LoRA ($p < 0.05$).}
  \label{tab:bootstrap_grouped}
  \begin{tabularx}{\textwidth}{@{}>{\raggedright\arraybackslash}Xlcc@{}}
   \toprule
\textbf{Scenario}      & \textbf{Metric} & $\mathbf{\Delta\text{ (RAG} - \text{RoBERTa-LoRA)}}$ & \textbf{95\% CI ($p$-value)}          \\ \midrule
\textbf{In-Domain}     & Precision       & $\mathbf{+0.0562}$                                   & $(+0.0271, +0.0873)^*$       \\
                       & Recall          & $-0.0614$                                   & $(-0.0897, -0.0326)^{**}$       \\
                       & F1-score        & $-0.0075$                                            & $(-0.0314, +0.0173)$ {[}$p = 0.55${]} \\ \midrule
\textbf{Hotel Reviews} & Precision       & $\mathbf{+0.1388}$                                   & $(+0.0715, +0.2071)^*$       \\
\textbf{(Fine-Tuned)}  & Recall          & $-0.2803$                                   & $(-0.3688, -0.194)^{**}$        \\
                       & F1-score        & ${+0.026}$                                             & $(-0.0456, +0.0913)$ {[}$p = 0.46${]} \\ \midrule
\textbf{Cross-Domain}  & Precision       & $\mathbf{+0.0416}$                                   & $(+0.0213, +0.0608)^*$       \\
\textbf{(Fine-Tuned)}  & Recall          & $-0.0455$                                   & $(-0.066, -0.0254)^{**}$        \\
                       & F1-score        & ${+0.0042}$                                            & $(-0.0115, +0.0195)$ {[}$p = 0.59${]} \\ \bottomrule
  \end{tabularx}
\end{table}
\FloatBarrier

In all three evaluation scenarios, paired hypothesis testing reveals a consistent and statistically significant trade-off between precision and recall.

The full RAG architecture provides a consistent and statistically significant increase in precision across all tests. In particular, precision increases by $+0.0562$ ($95\%$ CI: $[+0.0271, +0.0873]$) on the In-Domain scenario, by $+0.1388$ ($95\%$ CI: $[+0.0715, +0.2071]$) on the Fine-Tuned Hotel Reviews configuration, and by $+0.0416$ ($95\%$ CI: $[+0.0213, +0.0608]$) on the Fine-Tuned Cross-Domain case. This behaviour confirms that the combination of FAISS's vector search and the LLM's reasoning acts as a reliable filter for false positives. 

At the same time, the conservative verification of the LLM layer leads to a statistically significant reduction in recall across all configurations: $-0.0614$ ($95\%$ CI: $[-0.0897, -0.0326]$) for the In-Domain dataset, $-0.2803$ ($95\%$ CI: $[-0.3688, -0.1940]$) for hotel reviews, and $-0.0455$ ($95\%$ CI: $[-0.0660, -0.0254]$) for cross-domain testing. The most noticeable decrease is observed when there is a strong class imbalance (Hotel Reviews dataset). In this case, the LLM requires significantly more substantial textual evidence before predicting the ``fake'' label.  

The variation in the F1 metric remains statistically insignificant across all three configurations: $p=0.547$ for the In-Domain set, $p=0.461$ for the hotel reviews dataset, and $p=0.592$ for the cross-domain experiment. 

For completeness, the detailed paired bootstrap forest plots and numerical significance tables for the zero-shot baseline configurations (without parameter-efficient domain adaptation) are provided in ~\ref{sec:secA1_zeroshot}.

\subsubsection{Granular Error Analysis and Threshold Routing}
\begin{figure*}[!h]
    \centering
    \begin{subfigure}[b]{0.48\textwidth}
        \centering
        \includegraphics[width=\linewidth]{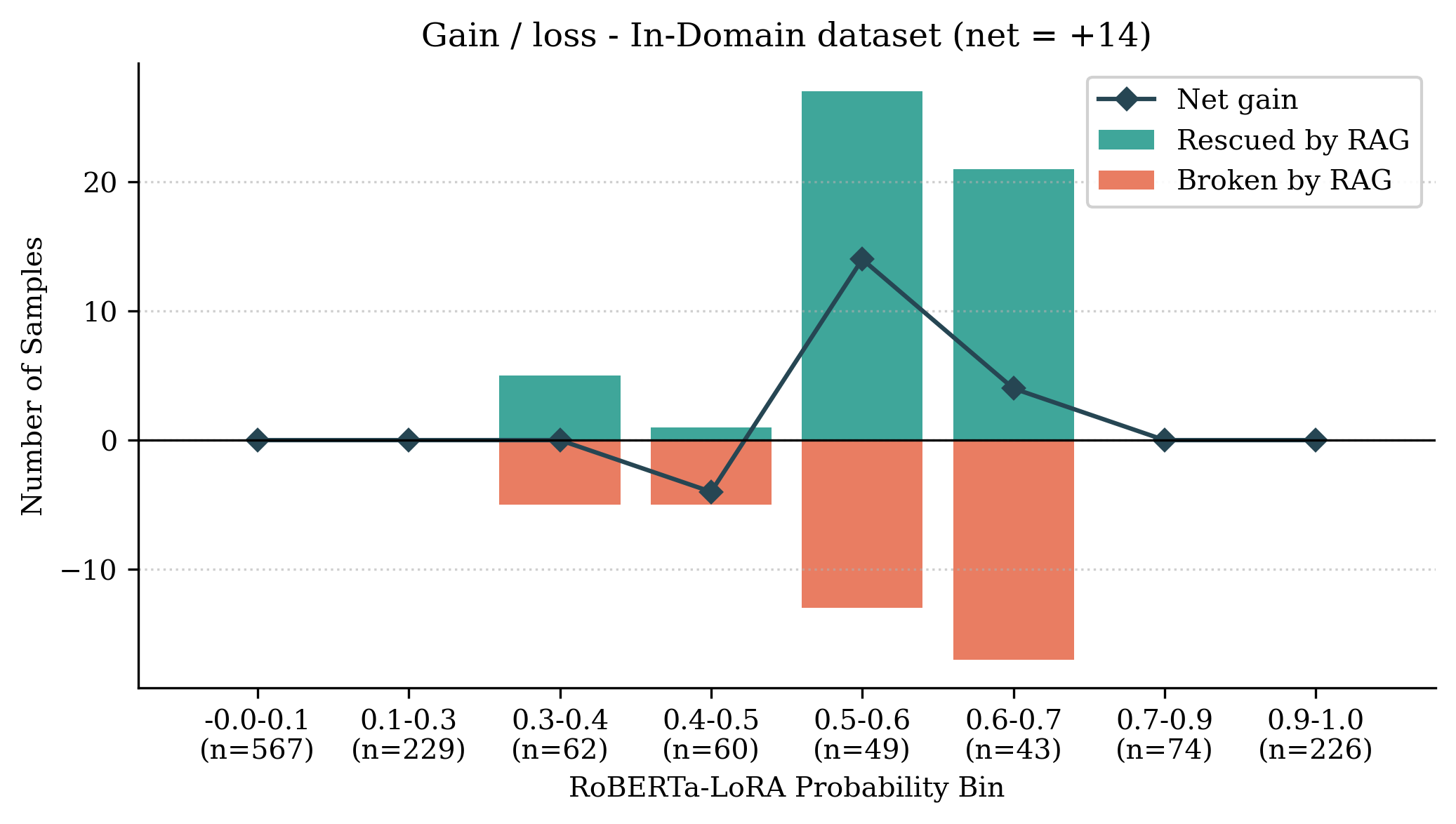}
    \end{subfigure}\hfill
    
    \vspace{0.8em}
    \begin{subfigure}[b]{0.48\textwidth}
        \centering
        \includegraphics[width=\linewidth]{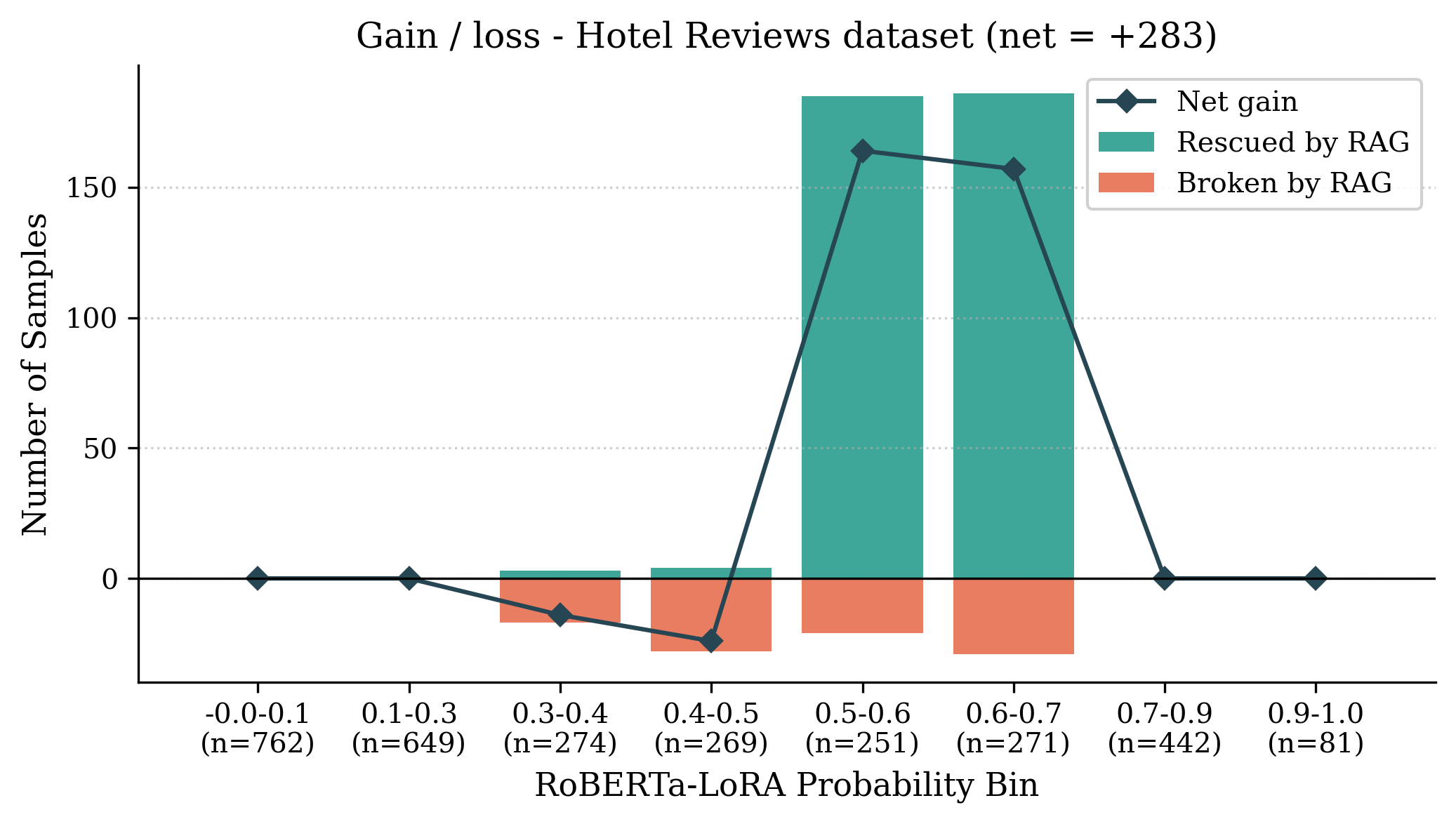}
    \end{subfigure}
    \begin{subfigure}[b]{0.48\textwidth}
        \centering
        \includegraphics[width=\linewidth]{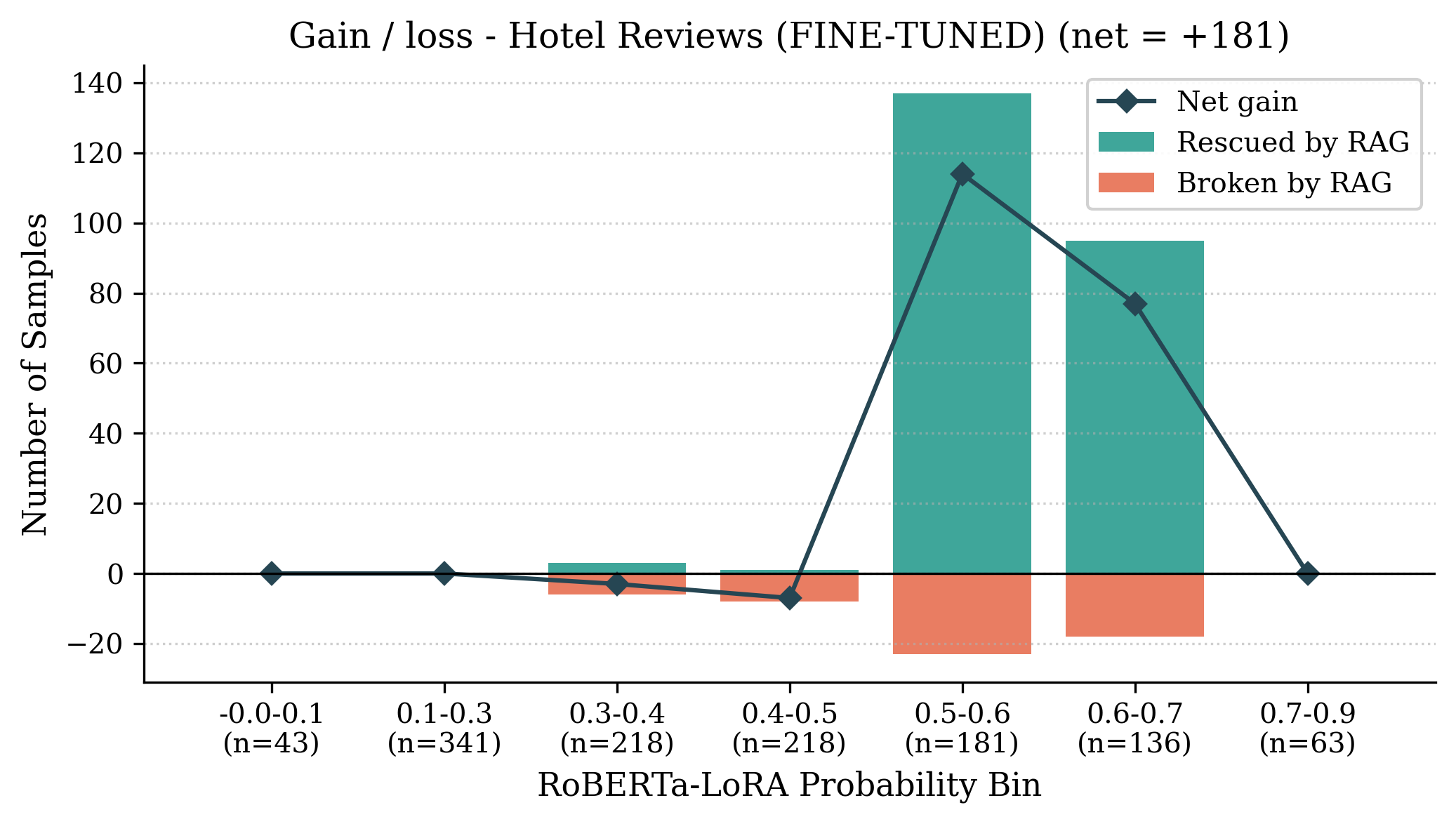}
    \end{subfigure}
    
    \vspace{0.8em}
    \begin{subfigure}[b]{0.48\textwidth}
        \centering
        \includegraphics[width=\linewidth]{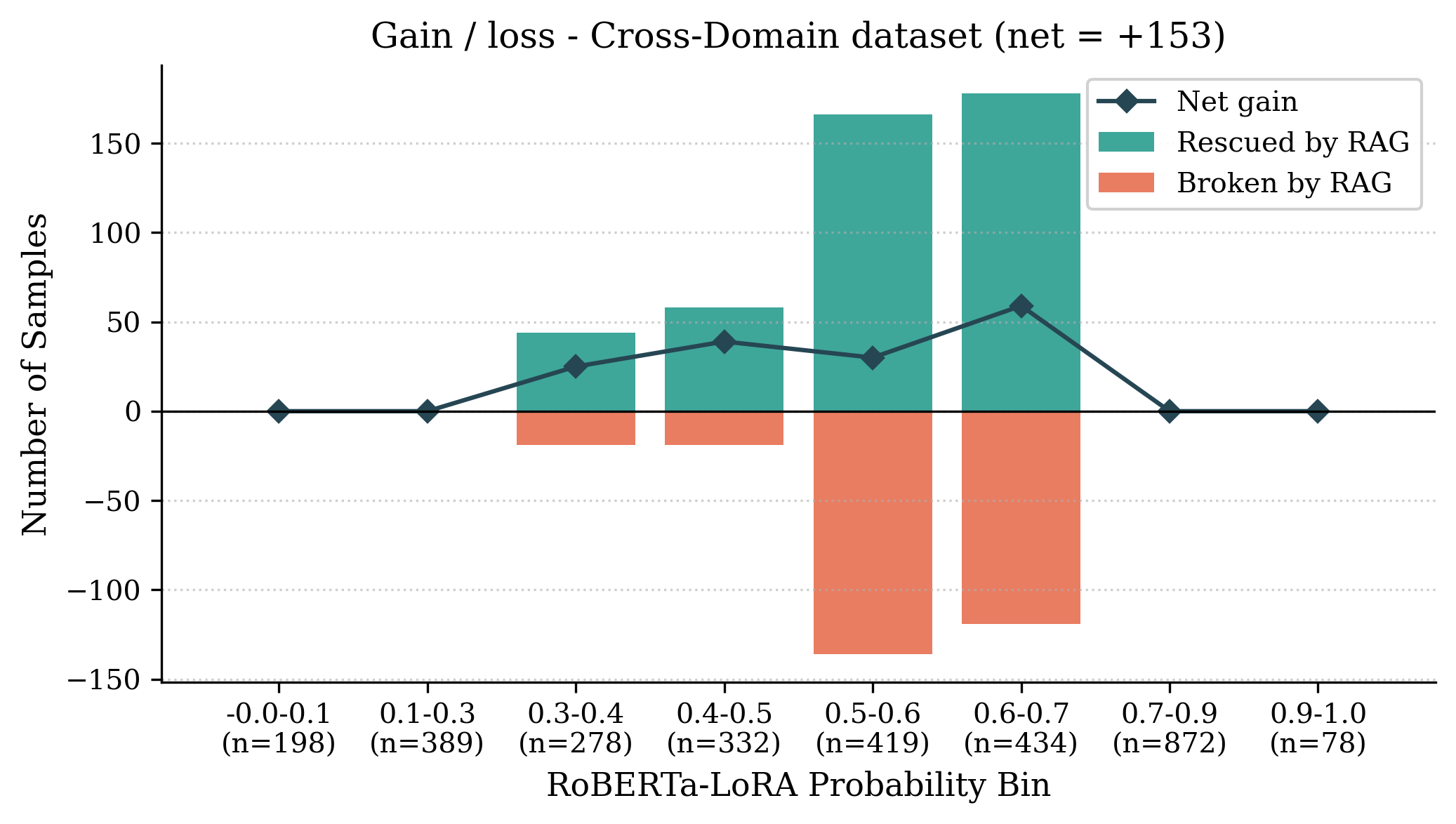}
    \end{subfigure}
    \begin{subfigure}[b]{0.48\textwidth}
        \centering
        \includegraphics[width=\linewidth]{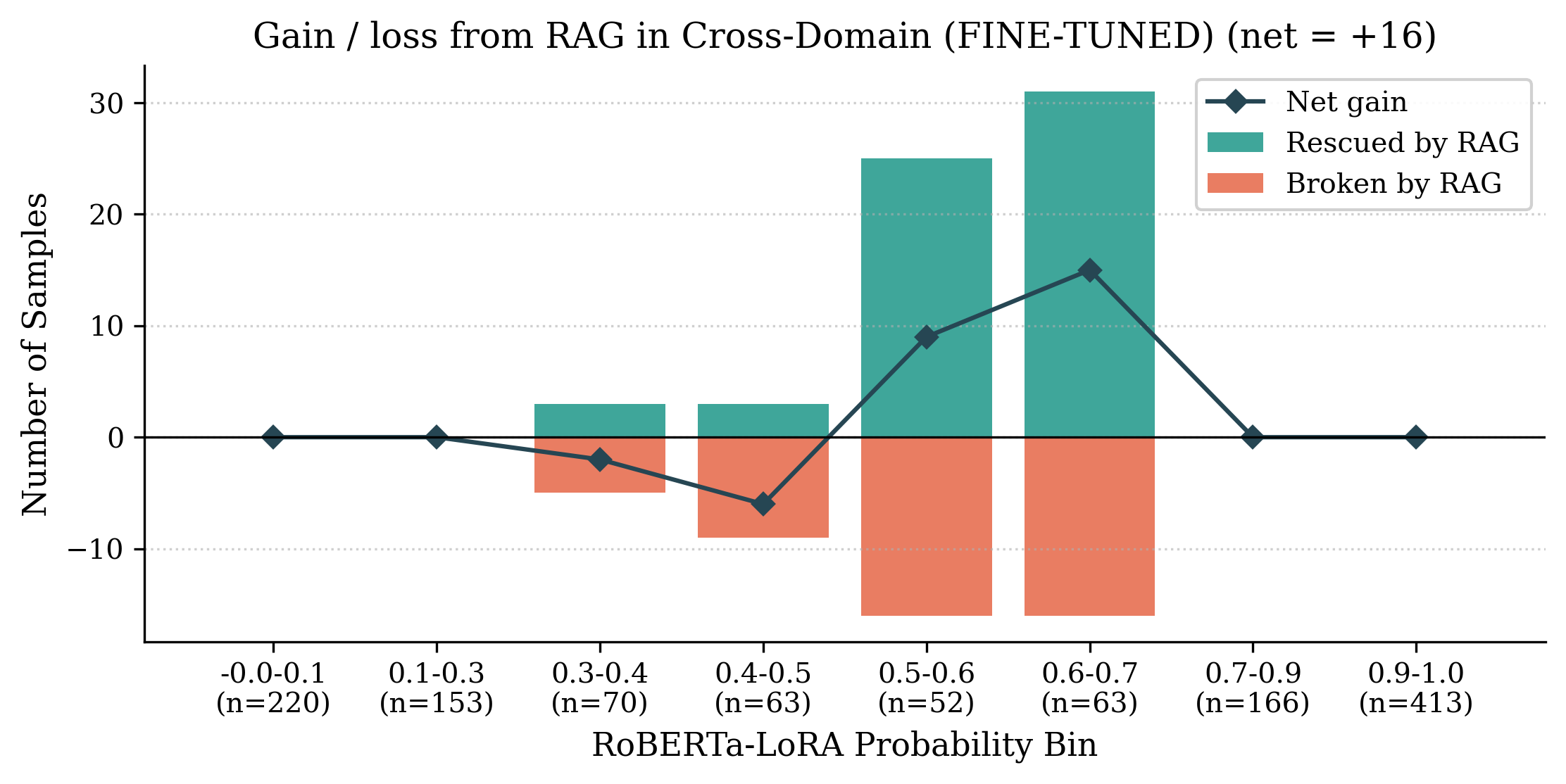}
    \end{subfigure}
    
    \caption{Effect of RAG on Model Predictions Across Confidence Bins across all experiments.}
    \label{fig:bins_gain_loss}
\end{figure*}

To evaluate the performance of the confidence-threshold routing mechanism, the decisions by the LLM layer were analysed across different probability confidence intervals of the primary RoBERTa-LoRA classifier. In Figure~\ref{fig:bins_gain_loss}, these corrections are classified as ``Rescued'' (incorrect base predictions successfully corrected by the RAG pipeline) and ``Broken'' (correct base predictions harmed by the LLM) across RoBERTa-LoRA confidence bins for all five experimental setups: In-Domain, Hotel Reviews (Zero-Shot and Fine-Tuned), and Cross-Domain AI-Spam (Zero-Shot and Fine-Tuned).

As expected from the routing rule in Section \ref{sec4}, the LLM made no changes
outside the uncertainty window ($p_{\text{base}} < 0.3$ and $p_{\text{base}} > 0.7$) in any
of the five setups. All corrections happen inside $[0.3, 0.7]$. Consequently, for unambiguous samples, it is possible to eliminate the GPU compute overhead and LLM latency.

The adjustments are strictly limited to the uncertainty window ($[0.3, 0.7]$). The highest positive net gains are consistently observed in the probability interval $0.5$–$0.7$ for all datasets. In this region, where the main classifier slightly favours a spam prediction with low confidence, the LLM, based on the FAISS retrieval evidence, acts as a high-precision verification filter, successfully correcting false alarms.

\FloatBarrier

Applying domain adaptation with effective parameter tuning radically changes the confidence distribution of the base classifier. In Zero-Shot settings and the Fine-Tuned Hotel Reviews configuration, the model’s high uncertainty causes a large number of samples to fall within the $[0.3, 0.7]$ interval (e.g., $n = 1065$ for Hotel Reviews Zero-Shot, $n = 763$ for Fine-Tuned Hotel Reviews, and $n = 1463$ for Cross-Domain Zero-Shot). In contrast, after fine-tuning in the cross-domain scenario, the number of uncertainty cases decreases significantly ($n = 248$ uncertain samples vs $n = 952$ samples with the model's high confidence). As a result, the RAG performs extensive corrections, leading to a significant net increase of $+283$ samples in the Zero-Shot Hotel Reviews configuration,  $+181$ in the Fine-Tuned Hotel Reviews setting, and $+153$ in the Zero-Shot Cross-Domain scenario. 

Fine-tuning in the Cross-Domain case increases the confidence level of the base classifier, shifting the majority of samples into extreme confidence intervals. Although this substantially reduces the number of samples entering the LLM pipeline, the RAG architecture maintains a consistently positive net correction rate.

\subsection{Comparison with Cross-Domain SOTA Baselines}
The performance of the proposed RAG system was compared with the baseline models established by Salminen et al. on the same cross-domain dataset to evaluate the robustness and competitiveness of the proposed RAG architecture. Their study provides benchmarks using an OpenAI GPT-2 output detector, NBSVM, and a fine-tuned fakeRoBERTa model \cite{Salminen2022}. 

The baseline models were trained and evaluated strictly  in-domain (using an 80/20 split of the target 40,432-review corpus). In contrast, the proposed RAG system was initially trained on a completely distinct domain (HoReCa reviews). To ensure strict evaluation fairness, the exact test split of $8,087$ reviews was reproduced using a fixed \texttt{random\_state = 2021}.

\begin{table}[htbp]
\centering
\small
\caption{Comparison of the proposed RAG system against baseline models on the cross-domain dataset.}
\label{tab:baseline_comparison}
\begin{tabular}{@{}>{\raggedright\arraybackslash}p{0.24\textwidth}>{\raggedright\arraybackslash}p{0.29\textwidth}ccc@{}}
\toprule
\textbf{Detector} & \textbf{Evaluation Mode} & \textbf{Precision} & \textbf{Recall} & \textbf{F1-Score} \\
\midrule
OpenAI Detector\textsuperscript{1} & In-domain (8k test set) & 0.83 & 0.82 & 0.82 \\
NBSVM\textsuperscript{1} & In-domain (8k test set) & 0.95 & 0.95 & 0.95 \\
fakeRoBERTa\textsuperscript{1} & In-domain (8k test set) & 0.97 & 0.97 & 0.97 \\
\midrule Proposed RAG System (No retraining) & zero-shot (8k test set) & 0.76 & 0.42 & 0.54 \\
Proposed RAG System  (With retraining) & 5 epochs adaptation \newline(8k test set) & 0.88 & 0.92 & 0.90 \\
\bottomrule
\end{tabular}
\par\vspace{4pt}
\begin{minipage}{\textwidth}
\footnotesize\raggedright
\textsuperscript{1}Note: Baseline metrics are extracted from Salminen et al. (2022) \cite{Salminen2022}.\par
\end{minipage}
\end{table}

Table \ref{tab:baseline_comparison} demonstrates the strong performance of the proposed architecture against target-domain baselines. Despite being initially trained on a substantially different domain, lightweight 5-epoch adaptation on a small sample of $2,264$ reviews enables our system to achieve a competitive F1-score of 0.90. Notably, the fine-tuned RAG system outperforms detection baselines such as the OpenAI Detector (F1 = $0.82$) and comes close to specialized in-domain models like fakeRoBERTa (F1 = $0.97$) and NBSVM (F1 = $0.95$). The proposed architecture achieves this level of accuracy using less than 7\% of the total training data for fine-tuning to a new domain ($2,264$ vs. $32,375$ reviews).

\subsection{Ablation Study} \label{subsec:ablation}
To evaluate the contribution of each component in the proposed architecture, an ablation study was conducted on a test sample from the in-domain dataset. During the experiments, four system configurations were compared sequentially: the baseline logistic regression model applied to RoBERTa embeddings (\textit{RoBERTa + LogReg}), the fine-tuned pretrained RoBERTa classifier using Low-Rank Adaptation (\textit{RoBERTa-LoRA}), the full RAG pipeline based on a RoBERTa-LoRA classifier with structured memory evidence and explicit vector metrics (\textit{Full RAG}), and the full RAG pipeline using a simplified prompt without explicit vector metrics and retrieved examples (\textit{Full RAG with Light Prompt}). 
The empirical results across all four configurations are summarised in Table~\ref{tab:ablation_rag}.

\begin{table}[h]
\centering
\small
\caption{Ablation analysis of pipeline components and prompting strategies on the in-domain test dataset.}
\label{tab:ablation_rag}
\begin{tabularx}{\textwidth}{@{}>{\raggedright\arraybackslash}Xcccc@{}}
\toprule
\textbf{Configuration} & \textbf{Accuracy} & \textbf{Precision} & \textbf{Recall} & \textbf{F1-Score}\\
\midrule
RoBERTa + LogReg & 0.7456 & 0.5590 & 0.6923 & 0.6186 \\
RoBERTa-LoRA & 0.8130 & 0.6862 & \textbf{0.6880} &  \textbf{0.6871} \\
Full RAG (light prompt)  & 0.8229 & 0.7331 & 0.6394 & 0.6831  \\
Full RAG & \textbf{0.8237} & \textbf{0.7424} & 0.6266  & 0.6796 \\
\bottomrule
\end{tabularx}
\end{table}
\FloatBarrier

Comparative analysis of these configurations reveals several insights:
\begin{itemize}
    \item \textbf{Baseline vs Parameter-Efficient Fine-Tuning}: Transitioning from a linear classifier built on RoBERTa embeddings to LoRA fine-tuning of the RoBERTa model significantly improves overall classification metrics, raising Accuracy from $0.7456$ to $0.8130$, Precision from $0.5590$ to $0.6862$, and F1-score from $0.6186$ to $0.6871$.
    \item \textbf{Effectiveness of Full RAG Reasoning}: Integrating the LLM decision layer with structured memory evidence to verify questionable reviews reduces false positives by increasing Precision from $0.6862$ to $0.7424$ while maintaining strong overall Accuracy ($0.8237$) compared with the RoBERTa-LoRA classifier.
    \item \textbf{Role of Explicit Similarity Metrics and Evidence}: Implementing a lightweight prompt that hides numerical FAISS vector distances and retrieved examples does not worsen the metrics, but interpretability — as will be shown in the following Section~\ref{sec:case_study} — decreases significantly.

\end{itemize}

To demonstrate the systemic weaknesses of the light prompt configuration, comprehensive cross-domain and imbalanced evaluations were added to~\ref{sec:secB_interolation_prompt} (Table~\ref{tab:ablation_add}). As detailed in~\ref{sec:secB_interolation_prompt}, using the light prompt under severe class imbalance (fine-tuned Hotel Reviews case) causes a sharp decline in the Precision metric from $0.3520$ to $0.1981$ and in F1-score from $0.3424$ to $0.2442$. In contrast, the Precision score improves across all evidence-backed settings. These findings confirm that providing structured numerical distance metrics and real evidence within a full RAG prompt is essential for model stability across diverse environments. 

\subsection{Qualitative Analysis \& Case Studies}
\label{sec:case_study}

Evaluating the transparency and interpretability of generated decision justifications is essential for deployment of an expert system in real-world content moderation workflows. To systematically measure the quality of LLM reasoning, the model's explanations generated within the classifier’s uncertainty window ($p_{\text{base}} \in [0.30; 0.70]$) were analysed. A comparative analysis of the proposed prompt for the RAG system, which includes explicit numerical FAISS metrics and examples from the database, against the light prompt, which omits explicit memory distances, was conducted. 

Four key explainability and reasoning performance dimensions were calculated:
\begin{enumerate}
    \item \textbf{Evidence Grounding Rate (EGR)}: The proportion of generated LLM explanations that have references to retrieved metrics and examples from the vector database (e.g., neighbour distance, fake ratio);
    \item \textbf{Average Quantitative Facts Count}: The average number of verifiable statistical parameters and vector metrics cited per explanation;
    \item \textbf{Positive Correction Rate (PCR)}: The frequency with which the LLM successfully overrides an incorrect baseline classifier prediction;
    \item \textbf{Mean Perplexity}: The evaluation of fluency and linguistic predictability of generated justifications using an external language model (Qwen-2.5 0.5B). While lower values indicate a more traditional text structure, local increases reflect the syntactic complexity associated with the inclusion of statistical data and specific probabilities in natural text.
\end{enumerate}

The empirical results across three setups (In-Domain dataset and fine-tuned configurations of experimental data) are presented in Table~\ref{tab:interpretability_results}.

\begin{table}[t]
\centering
\scriptsize
\setlength{\tabcolsep}{4pt}
\caption{Grounding and reasoning metrics within the uncertainty window ($p_{\text{base}} \in [0.30, 0.70]$).}
\begin{tabularx}{\textwidth}{@{}>{\raggedright\arraybackslash}Xlcccc@{}}
\toprule
\textbf{Scenario} & \textbf{Setup} & \textbf{EGR (\%)}  & \textbf{Avg. Facts Count} & \textbf{PCR (\%)}  & \textbf{Perplexity} \\
\midrule
{\textit{In-Domain Baseline}} 
& Light Prompt & 0.5          & 1.01   & \textbf{25.6}        & \textbf{32.89}  \\
& Full RAG  & \textbf{100.0} & \textbf{3.50} & 25.1 & 41.07 \\

\midrule
{\textit{Hotel Reviews}}
& Light Prompt & 0.5          & 1.04   &   23.1    & 30.61  \\
{\textit{(Fine-Tuned)}} & Full RAG  & \textbf{99.5}  & \textbf{6.32} & \textbf{31.3} & \textbf{27.55}  \\

\midrule
{\textit{Cross-Domain }} 
& Light Prompt & 0.0           & 1.08   &   19.3    & \textbf{27.61}  \\
{\textit{(Fine-Tuned)}} & Full RAG  & \textbf{80.5}  & \textbf{1.98} & \textbf{25.0} & 40.40  \\

\bottomrule
\label{tab:interpretability_results}
\end{tabularx}
\end{table}
\FloatBarrier

As shown in Table~\ref{tab:interpretability_results}, hiding the FAISS numerical distances in the configuration with the light prompt obviously leads to an almost complete collapse of grounding, reducing the EGR metric to 0.0-0.5 in all scenarios. In contrast, the system with the proposed prompt achieves nearly perfect grounding justification ($99.5\%$–$100.0\%$ in in-domain and hotel reviews configurations, as well as $80.5\%$ in cross-domain settings), consistently integrating an average of $1.98$ to $6.32$ statistical parameters into each decision. 

Furthermore, the positive correction rate (PCR) demonstrates that the \textit{Full RAG} pipeline successfully corrects between $25.0\%$ and $31.3\%$ of baseline classification errors. Under challenging conditions, such as an imbalanced hotel reviews dataset and a cross-domain task, the proposed prompt achieves a significantly higher rate of positive correction than the \textit{Light Prompt} variant ($31.3\%$ vs $23.1\%$ in hotel reviews, $25.0\%$ vs $19.3\%$ in a cross-domain environment).

With respect to the linguistic structure of LLM explanations, the Perplexity metric, estimated using  Qwen-2.5 (0.5B), provides important information about the density of justifications. In the In-Domain ($41.07$ vs $32.89$) and Cross-Domain ($40.40$ vs $27.61$) configurations, the \textit{Full RAG Prompt} exhibits significantly higher perplexity than the \textit{Light Prompt}. This increase in perplexity does not indicate poor readability, but is directly due to the dense inclusion of non-standard numerical constructs, such as exact floating-point vector distances ($0.1082$), similarity thresholds, and specific classifier probabilities ($p_{\text{base}} = 0.615$). These numerical tokens introduce local syntactic complexity that differs from standard conversational pre-training text, which reduces predictability for the Qwen evaluator. Curiously, in the imbalanced Hotel Reviews dataset, the justifications generated using \textit{Full RAG} exhibit both an exceptionally high fact density ($\text{Avg. Facts Count} = 6.32$) and a lower perplexity score ($27.55$), indicating that specialised fine-tuning allows the LLM to seamlessly integrate complex numerical evidence without sacrificing natural sentence flow.

To show how evidence-based reasoning eliminates classification errors in borderline cases, Table~\ref{tab:case_study_rescue} qualitatively compares the explanations generated for ambiguous cases.

\begin{table*}[!t]
\centering
\footnotesize
\linespread{0.9}\selectfont
\renewcommand{\arraystretch}{0.9}
\setlength{\tabcolsep}{4pt}
\caption{Side-by-side comparison of generated explanations on two uncertain test instances. Bold highlights verifiable statistical evidence; italicised text indicates ungrounded speculation.}
\label{tab:case_study_rescue}
\begin{tabular}{@{}p{0.20\textwidth}p{0.38\textwidth}p{0.38\textwidth}@{}}
\toprule
\textbf{Input Review \& Metadata} & \textbf{Light Prompt (Ablation)} & \textbf{Full RAG Prompt (Ours)} \\
\midrule
\textit{``GREAT CORN Came here on a late Monday night and was one of the last customers I was very pleased that the waitress did not try to rush us out''} \vspace{0.4em} \newline
\textbf{Base Probability:} \newline
$p_{\text{base}} = 0.61$ \newline
(Uncertain) \newline
\textbf{Retrieved Stat:} \newline
$\text{fake\_ratio} = 0.10$ \newline
$2/3$ examples - reals  \newline
\textbf{Ground Truth:} \newline
$y = 0$ (Genuine)
& 
\textbf{Prediction:} $1$ (Deceptive) $\times$ \newline
\textbf{Output:} \newline
\textit{``LABEL: 1. REASONING: The review contains generic and templated text (``GREAT CORN'') and lacks situational context, making it seem like a pre-written promotional phrase. The mention of a specific day and time (``late Monday night'') is not enough to ground the review in reality.''}
& 
\textbf{Prediction:} $0$ (Genuine) $\checkmark$ \newline
\textbf{Output:} \newline
\textit{``LABEL: 0. REASONING: The retrieved examples are mostly real ($2$ out of $3$), with a fake ratio of $0.1$ in the top-20 neighbours. The closest distance is $0.1082$, indicating a moderate similarity to memory. The classifier result is $0.615$, which is not conclusive. The text is grounded and situational, describing a specific experience at a restaurant. The mention of a late Monday night and a specific interaction with the waitress suggests a genuine review. The use of ``GREAT CORN'' in all caps may be a minor red flag, but it is not enough to override the other evidence.''} \\
\midrule
\textit{``I feel like i need to go back and read this book again I felt it was very repetitive and not applicable to most Id love to be able to quit work move to the beach and relax That is simply not an option''} \vspace{0.4em} \newline
\textbf{Base Probability:} \newline
$p_{\text{base}} = 0.67$ \newline
(Uncertain) \newline
\textbf{Retrieved Stat:} \newline
$\text{fake\_ratio} = 0.00$ \newline
$3/3$ examples - reals  \newline
\textbf{Ground Truth:} \newline
$y = 0$ (Genuine)
& 
\textbf{Prediction:} $1$ (Deceptive) $\times$ \newline
\textbf{Output:} \newline
\textit{``LABEL: 1 REASONING: The review contains generic and templated language, such as ``I feel like I need to go back and read this book again'' and ``Id love to be able to quit work move to the beach and relax'', which are common phrases used in promotional or spam reviews. The text also lacks specific details about the book, making it seem like a generic complaint rather than a genuine review.''}
& 
\textbf{Prediction:} $0$ (Genuine) $\checkmark$ \newline
\textbf{Output:} \newline
\textit{``LABEL: 0 REASONING: The retrieved examples are mostly real ($3/3$), with a fake ratio of $0.00$ in the top-20 neighbours. The closest distance is $0.0937$, indicating a moderate similarity to memory. The classifier result is a probability of spam ($0.675$), but I prioritise the retrieval evidence over the classifier result. The target review text is grounded and situational, expressing a personal opinion about a book. The language is imperfect and experience-based, describing a feeling of repetition and inapplicability. The tone is not overly promotional or templated, and the text does not exhibit generic or repetitive language.''} \\
\bottomrule
\end{tabular}
\end{table*}
\FloatBarrier

The qualitative comparisons presented in Table~\ref{tab:case_study_rescue} illustrate the fundamental difference between ungrounded generative reasoning (\textit{Light Prompt}) and vector-based reasoning (\textit{Full RAG Prompt}):

\begin{enumerate}
    \item \textbf{Vulnerability to stylistic and structural artifacts}: The LLM provided with a simplified prompt penalises stylistic informalities, specifically, capitalised headings (``GREAT CORN'') or emotional and colloquial expressions (``I'd love to be able to quit work''), mistakenly labelling them as formulaic deception. This results in false positives, where genuine, situational customer feedback is wrongly penalised for stylistic informalities.
    \item \textbf{Retrieval Neighborhood Geometry for Evidence Grounding}: The \textit{classical RAG prompt} provides explicit quantitative memory signals ($\mathit{fake\_ratio} = 0.10$ and $0.00$; nearest distances $0.1082$ and $0.0937$). These numerical metrics inform the LLM that the target review is within a dense and statistically homogeneous cluster of verified genuine reviews. Guided by this spatial evidence, the LLM correctly recognises that minor stylistic noise does not override strong memory proximity.
    \item \textbf{Overcoming Baseline Classifier Bias}: In borderline cases, when the primary classifier outputs moderate spam probabilities ($p_{\text{base}} = 0.61$ and $0.67$), ungrounded LLMs tend to confirm the classifier bias toward fraud. In contrast, the structured RAG prompt lets the generative layer actively overcome the classifier’s uncertainty whenever the retrieved vector memory provides overwhelming evidence of authenticity, directly preventing false positives.
\end{enumerate}

\section{Discussion} \label{sec6}
An empirical analysis of the experimental results demonstrates the performance capabilities, structural trade-offs, and generalisation limits of the hybrid Retrieval-Augmented Generation (RAG) architecture. This section presents a comprehensive analysis of the results, which directly addresses the main research questions (RQ1–RQ4).

\subsection{Performance Improvement via Retrieval-Augmented Reasoning (RQ1)}
The experimental data confirm that combining the RoBERTa-LoRA classifier with FAISS vector search and LLM-based reasoning shifts the classifier toward a higher-precision operating point while maintaining statistically competitive overall detection performance. Across all evaluated scenarios, the full RAG pipeline achieves a statistically significant increase in Precision: $+0.0562$ on the In-Domain dataset (rising from $0.6862$ to $0.7424$, $\text{p-value} < 0.05$), $+0.1388$ on the imbalanced Hotel Reviews dataset (rising from $0.2132$ to $0.3520$, $\text{p-value} < 0.05$), and $+0.0416$ on the Cross-Domain dataset (rising from $0.7651$ to $0.8067$, $\text{p-value} < 0.05$).

This increase in Precision results in a statistically significant decrease in Recall ($-0.0614$ In-Domain, $-0.2803$ Hotel Reviews, $-0.0455$ Cross-Domain), whereas changes in the overall F1-score remain statistically insignificant ($\text{p-value} > 0.05$). In practical content moderation, this conservative verification threshold is highly advantageous, because falsely labelling a genuine user review as fake causes legitimate sellers greater commercial and reputational damage than failing to detect an occasional fraudulent review. Furthermore, the threshold routing mechanism ($\tau_{low} = 0.3$, $\tau_{high} = 0.7$) restricts the LLM outputs exclusively to borderline samples within the uncertainty interval ($p_{base} \in [0.3; 0.7]$), yielding a net reduction in classification errors (+14 within the domain, +181 for fine-tuned hotel reviews, +16 for fine-tuned cross-domain reviews). Moreover, this approach completely avoids generative compute overhead for deterministic predictions, leading to substantial reductions in energy consumption and the operational carbon footprint.

\subsection{Cross-Domain Adaptation Efficiency via LoRA (RQ2)}
Evaluating the system under zero-shot conditions on unfamiliar domains yields modest performance (Cross-Domain Accuracy $0.6300$, F1-score $0.6075$) due to domain shift and new AI-generated linguistic structures. However, implementing parameter-efficient fine-tuning using LoRA adapters in just 5 additional epochs enables fast domain transfer, raising Cross-Domain Accuracy to $0.8433$, Recall to $0.8829$, and F1-score to $0.8431$.

When evaluated against target-domain baselines established by Salminen et al. (OpenAI Detector F1 = 0.82, NBSVM F1 = 0.95, fakeRoBERTa F1 = 0.97), which were trained exclusively in-domain, the proposed RAG system achieves competitive accuracy (F1 = 0.90, Precision = 0.88, Recall = 0.92) despite initial pre-training on an entirely different corpus of hotel and restaurant reviews. It is important to note that the proposed methodology achieves these results using less than 7\% of the training data to adapt to the target domain ($2,264$ reviews compared to the $32,375$ used in the SOTA models within a single domain), which demonstrates both highly efficient data utilization and strong cross-domain transferability. Additionally, adaptation using LoRA improves the reliability of the base classifier predictions by shifting the sample distributions toward high-confidence ranges and reducing the number of ambiguous reviews requiring LLM evaluation, specifically, from 43\% (zero-shot) of unclear cases to 12\% (after fine-tuning), consequently optimising computational throughput during inference. 

\subsection{Architecture Robustness Under Class Imbalance (RQ3)}
Review distributions in e-commerce are often characterised by extreme class imbalance similar to the Hotel Reviews scenario with 90\% of genuine and 10\% of fake reviews. Traditional classifiers using the standard cross-entropy loss function suffer from a strong bias toward the majority class, as demonstrated by the TF-IDF + Naive Bayes baseline, which achieved a Recall of only $0.0897$. To mitigate this problem, the base RoBERTa-LoRA classifier incorporates a dynamically weighted cross-entropy loss function that penalises minority class misclassifications approximately 9 times more heavily in the case of a 90\%/10\% imbalance.  

Under a 90\%/10\% class skew, applying 5 epochs of LoRA adaptation elevates overall Accuracy from $0.7083$ to $0.8592$, Precision from $0.2132$ to $0.3520$, and F1-score from $0.3164$ to $0.3424$. The LLM generative layer in the full RAG architecture maintains a conservative decision threshold under high class imbalance, requiring stronger evidence from the vector knowledge base. Although it results in a decrease in Recall, it provides a significant improvement in Precision ($+0.1388$, $\text{p-value} < 0.05$) and a net gain of $+181$ correctly reclassified reviews within the uncertainty window. 

\subsection{Interpretability Without Performance Degradation (RQ4)}
Standard transformer-based classifiers operate as “black-box” models and do not provide verifiable justifications for their predictions. Ablation analysis demonstrates that integrating LLM reasoning with FAISS vector memory allows us to maintain baseline classification quality (Accuracy $0.8237$ vs $0.8130$, F1-score $0.6796$ vs $0.6871$ on In-Domain data) while ensuring transparency in decision-making. 

It is important to note that providing explicit numerical vector metrics (nearest neighbour distance, fake ratio among the 20 nearest neighbours) and the top-3 examples from within the proposed RAG prompt yields an Evidence Grounding Rate (EGR) of $99.5\%$–$100.0\%$ in In-Domain and Hotel Reviews settings, providing an average of 1.98 to 6.32 statistical metrics per explanation. In contrast, removing explicit distance metrics (Light prompt experiment) obviously causes the EGR to drop to 0.0–0.5\%, triggering LLM hallucinations in which harmless stylistic features (such as capital letters in headings or emotional phrasing) are mistakenly interpreted as deceptive indicators. The spatial neighbourhood geometry provided by FAISS allows the LLM to recognise when a target review falls within a dense cluster of verified reviews, overriding base-classifier uncertainty and reducing false alarms.

\subsection{Practical Implications and Limitations }

From an operational perspective, the proposed hybrid RAG architecture establishes a balanced, scalable framework for real-world automated moderation platforms through three integrated capabilities. First, lightweight 5-epoch LoRA fine-tuning enables rapid domain adaptation; when expanding to new product categories (e.g., transitioning from hospitality to consumer electronics), platform administrators can update only the low-rank adapter using a minimal labelled sample, achieving substantial performance gains without retraining the backbone network. Second, the threshold-gated routing mechanism optimises inference latency and computational overhead, bypassing LLM invocation when classifier confidence is high and thereby reducing operational carbon footprint. Third, evidence-grounded reasoning provides trust-and-safety teams with explicit, auditable decision traces retrieved directly from vector memory.

Ultimately, this work demonstrates that Retrieval-Augmented Generation transforms deceptive review detection from opaque binary classification into an auditable expert content moderation system. Rather than serving purely as a standalone predictor, the RAG architecture operates as an expert auditing engine that reconciles high precision with verifiable textual evidence. For platform administrators, trust-and-safety teams, and regulatory compliance auditors, this transparent reasoning trail bridges the gap between automated deep learning predictions and defensible human-in-the-loop decision-making in real-world commercial environments.

Despite these strengths, the system has several limitations. First, under severe class imbalance (90/10), the fine-tuned Full RAG pipeline reaches a precision of $0.3520$ and a recall of $0.3333$. This means that about two thirds of the classified reviews are genuine and about two thirds of the fake reviews are missed, so the system should not work alone in such conditions. Second, the LLM layer provides the classifier with a higher precision and a lower recall, while the F1-score does not change significantly. We did not test whether a simple change of the classifier decision threshold gives the same effect, so this research is still open. Third, explainability was measured with automatic metrics only. The Evidence Grounding Rate shows that explanations cite the retrieved evidence, but not that they use it correctly, and we did not run a study with human moderators. Finally, all results come from a single run with one LLM (Llama-3.1-8B, 8-bit), fixed thresholds, and labelled target-domain data for the 5-epoch adaptation.

\section{Conclusions and future work} \label{sec7}

This paper presents a hybrid architecture based on RAG designed for the automatic detection and explanation of deceptive online reviews. By structurally integrating a supervised RoBERTa-LoRA classifier, a FAISS vector search module, and a large language model, the proposed system overcomes the fundamental limitations of traditional “black-box” neural network classifiers and traditional lexical models. The main contribution of this research is transforming deceptive review detection into an evidence-grounded, auditable decision-making pipeline in which each prediction is accompanied by a human-understandable justification based on evidence retrieved from vector memory.

The experimental evaluation across three different scenarios (in-domain, a highly asymmetric distribution of $90\%$/$10\%$ of hotel reviews, and a balanced cross-domain AI-generated spam) demonstrates the robustness and generalisation ability of the proposed architecture. The full RAG acts as a high-precision verification filter, delivering statistically significant increases in Precision across all evaluated setups ($+0.0562$ In-Domain, $+0.1388$ Imbalanced data, $+0.0416$ Cross-Domain, $\text{p-value} < 0.05$). Lightweight LoRA fine-tuning in just 5 additional epochs enables rapid cross-domain transfer, elevating F1-score to 0.90 on benchmark evaluation and outperforming OpenAI Detector (F1 = $0.82$) despite using minimal data for adaptation process. To optimise efficiency, the confidence threshold routing mechanism strictly limits LLM inference to borderline predictions, significantly reducing computational overhead. Furthermore, providing explicit numerical vector metrics and retrieved examples achieves an Evidence Grounding Rate of $80.5\%$–$100.0\%$, effectively eliminating hallucinations. Although integrating the retrieval exact numerical scores increases local perplexity ($27.55$–$41.07$), this reflects the structural density of quantitative evidence rather than degraded readability.

Future work will focus on several key directions. First, we plan to analyse trade-offs between precision–recall trade-offs under different operational settings and investigate deployment in agentic moderation environments. To address current operational constraints, future extensions will explore minority class resampling techniques (such as SMOTE and ADASYN), expand the vector knowledge base across broader commercial domains, integrate higher-capacity LLM backbones, and extend the framework to multilingual review environments.

\section*{Acknowledgments}
\label{Acknowledgments}
The research reported in this paper was supported by the European Union NextGenerationEU/PRTR, and FederaTrans project: Grant PID2024-158373OB-I00 funded by MICIU/AEI/10.13039/501100011033 and by ERDF/EU. Finally, the research reported in this paper is also funded by the European Union (CUSTOMAI project, grant agreement no. 101226029).

\appendix
\section{Zero-Shot Significance Analysis}\label{sec:secA1_zeroshot}

\begin{figure*}[!h]
    \centering
    \begin{subfigure}[b]{0.48\textwidth}
        \centering
        \includegraphics[width=\linewidth]{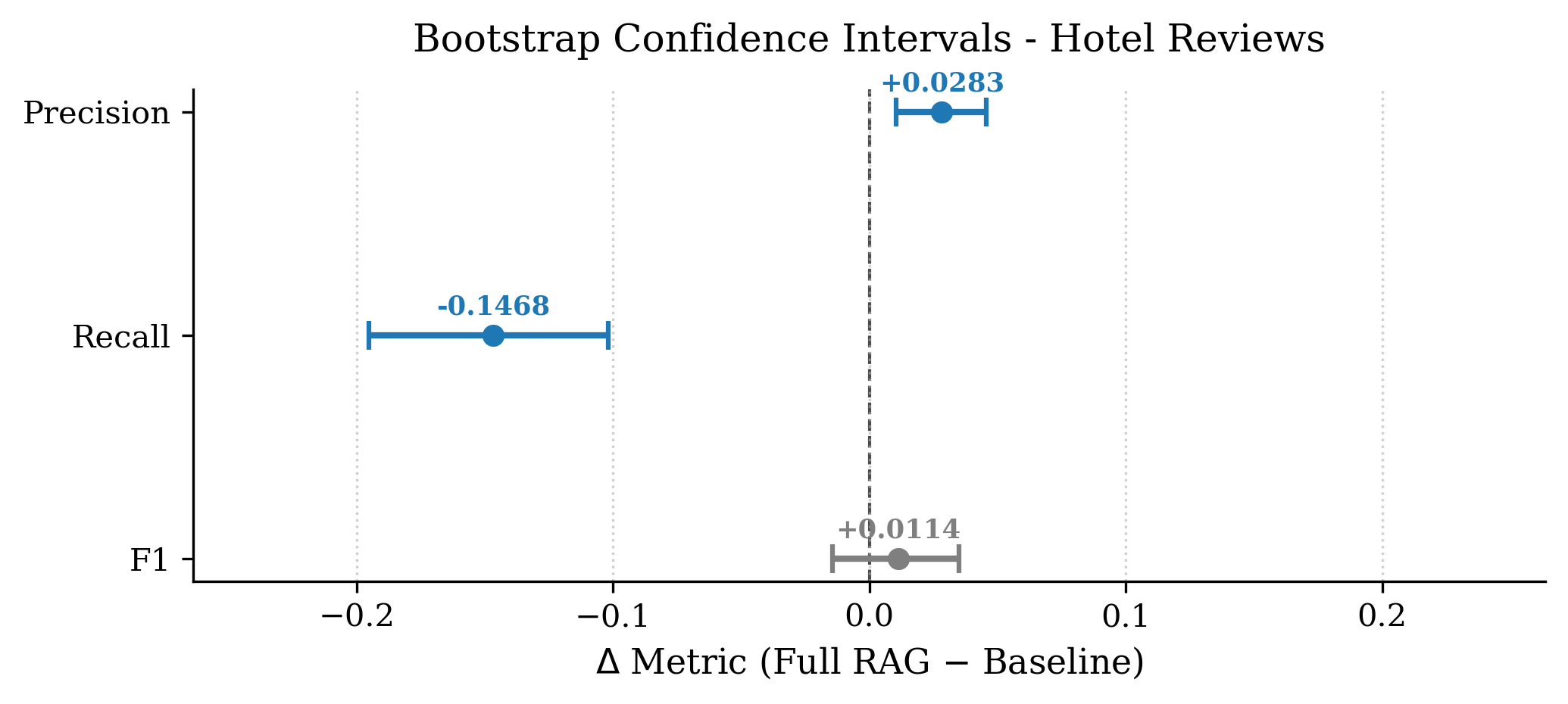}
    \end{subfigure}
    \begin{subfigure}[b]{0.48\textwidth}
        \centering
        \includegraphics[width=\linewidth]{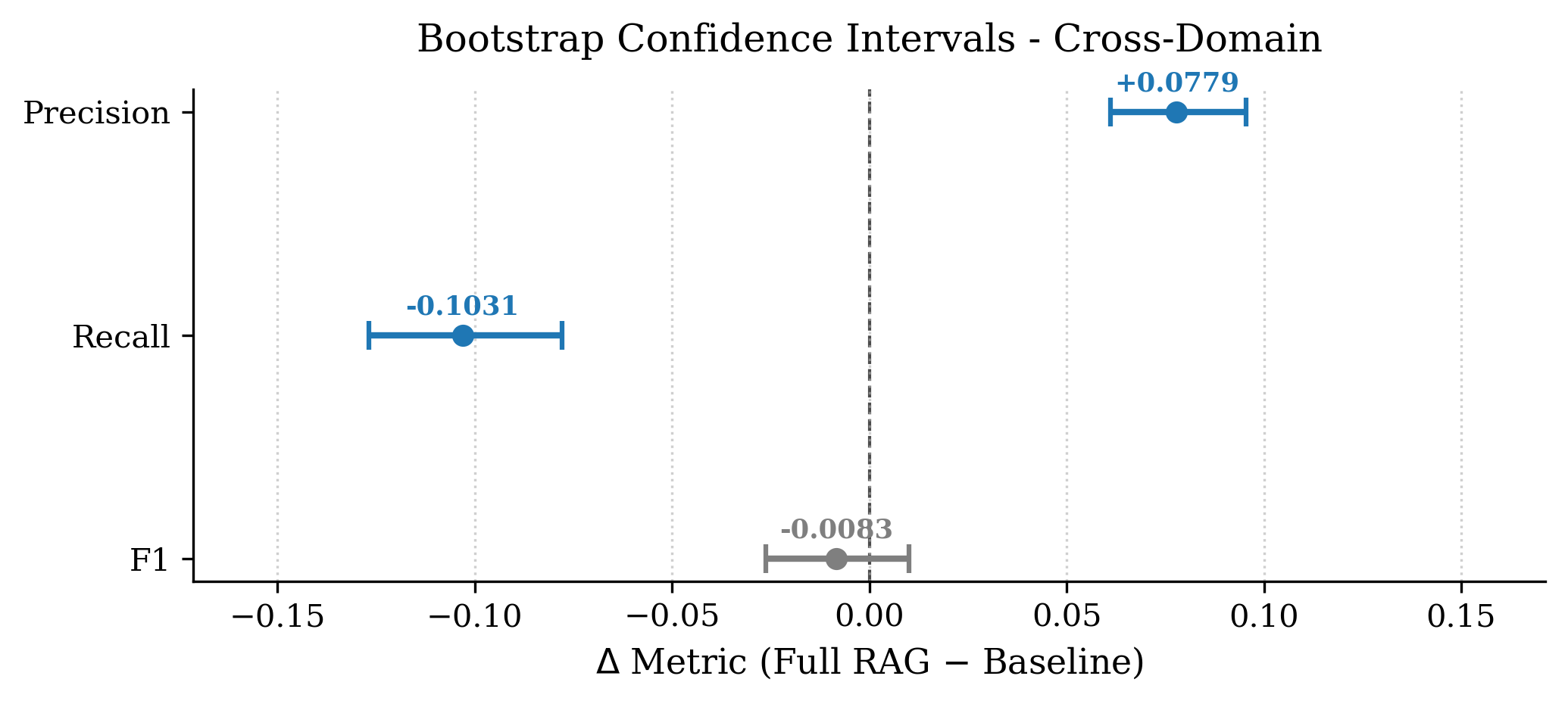}
    \end{subfigure}
    
    \caption{Bootstrap confidence intervals ($95\%$ CI) between Full RAG and Baseline across zero-shot configurations.}
    \label{fig:bootstrap_zerohot}
\end{figure*}

To complement the main evaluation, this appendix presents the detailed paired bootstrap significance analysis for the zero-shot baseline setups. Figure~\ref{fig:bootstrap_zerohot} illustrates the bootstrap difference distributions, and Table~\ref{tab:bootstrap_zeroshot} details the precise numerical metrics and confidence intervals.

\begin{table}[h]
  \centering
  \small
  \caption{Paired Bootstrap Significance Analysis ($N = 2{,}000$). Asterisks ($^*$) denote statistically significant differences in favour of our proposal, while double asterisks ($^{**}$) mean significant differences in favour of RoBERTa-LoRA ($p < 0.05$).}
  \label{tab:bootstrap_zeroshot}
  \begin{tabularx}{\textwidth}{@{}>{\raggedright\arraybackslash}Xlcc@{}}
    \toprule
    \textbf{Scenario} & \textbf{Metric} & $\mathbf{\Delta\text{ (RAG} - \text{Base)}}$ & \textbf{95\% CI ($p$-value)} \\
    \midrule
    \textbf{Hotel Reviews (zero-shot)} 
      & Precision & $\mathbf{+0.0283}$ & $(+0.0103, +0.0457)^*$   \\
      & Recall    & $-0.1468$          & $(-0.1953, -0.1018)^{**}$ \\
      & F1-Score  & $+0.0114$ & $(-0.0145, +0.035)$ [$p = .341$]  \\
    \midrule
    \textbf{Cross-Domain (zero-shot)} 
      & Precision & $\mathbf{+0.0779}$ & $(+0.06113, +0.0954)^*$   \\
      & Recall    & $-0.1031$ & $(-0.1269, -0.0778)^{**}$   \\
      & F1-Score  & $-0.0083$          & $(-0.0262, +0.001)$ [$p = .388$] \\
    \bottomrule
  \end{tabularx}
\end{table}
\FloatBarrier

\FloatBarrier

\section{Additional Experiments with Light Prompt}
\label{sec:secB_interolation_prompt}
\begin{table}[htbp]
\centering
\footnotesize
\setlength{\tabcolsep}{2pt}
\caption{Ablation study of pipeline components and prompting strategies on the Hotel Reviews and Cross-Domain scenarios.}
\label{tab:ablation_add}
\begin{tabularx}{\textwidth}{@{}l>{\raggedright\arraybackslash}Xcccc@{}}
\toprule
\textbf{Dataset} & \textbf{Setup} & \textbf{Acc}  & \textbf{Prec} & \textbf{Rec} & \textbf{F1-Score} \\
\midrule
\multicolumn{6}{l}{\textit{Hotel Reviews (zero-shot)}} \\
& RoBERTa-LoRA & 0.6712 & 0.1684 & \textbf{0.6007} & 0.2631   \\
& Full RAG (light prompt) & 0.7456 & 0.1807 & 0.4539 & 0.2585  \\
& Full RAG & \textbf{0.7656} & \textbf{0.1967} & 0.4539 & \textbf{0.2745}  \\
\midrule
\multicolumn{6}{l}{\textit{Hotel Reviews (fine-tuned)}} \\
& RoBERTa-LoRA  & 0.7083 & 0.2132 & \textbf{0.6136} & 0.3164  \\
& Full RAG (light prompt) & 0.7833 & 0.1981 & 0.3182 & 0.2442 \\
& Full RAG & \textbf{0.8592} & \textbf{0.3520} & 0.3333 & \textbf{0.3424} \\
\midrule
\multicolumn{6}{l}{\textit{Cross-Domain (zero-shot)}} \\
& RoBERTa-LoRA  & 0.5790 & 0.5613 & \textbf{0.6819} & 0.6158  \\
& Full RAG (light prompt) & 0.6270 & 0.6156 & 0.6550 & \textbf{0.6347} \\
& RAG (RoBERTa-LoRA) & \textbf{0.6300} & \textbf{0.6391} & 0.5788 & 0.6075 \\
\midrule
\multicolumn{6}{l}{\textit{Cross-Domain (fine-tuned)}} \\
& RoBERTa-LoRA  & 0.8300 & 0.7651 & {0.9283} & 0.8389  \\
& Full RAG (light prompt) & 0.8292 & 0.7603 & \textbf{0.9371} & 0.8395  \\
& Full RAG & \textbf{0.8433} & \textbf{0.8067} & 0.8829 & \textbf{0.8431}  \\
\bottomrule
\end{tabularx}
\end{table}
\FloatBarrier
% ...

\end{document}